\documentclass{article}

     \PassOptionsToPackage{numbers, sort&compress}{natbib}

\usepackage[final,main]{neurips_2026}

\usepackage[utf8]{inputenc} 
\usepackage[T1]{fontenc}    
\usepackage{hyperref}       
\usepackage{url}            
\usepackage{booktabs}       
\usepackage{amsfonts}       
\usepackage{nicefrac}       
\usepackage{microtype}      
\usepackage{xcolor}         
\usepackage{graphicx}
\usepackage{amsmath}
\usepackage{wrapfig}
\usepackage{colortbl} 
\usepackage{subcaption}
\usepackage{graphicx}
\usepackage{multirow}
\usepackage{pifont}
\definecolor{mygray}{gray}{.9}
\definecolor{mygreen}{HTML}{27AE60}
\definecolor{myred}{HTML}{E74C3C}

\newcommand{\gup}[1]{\ensuremath{_{{\color{mygreen}\uparrow#1}}}}
\newcommand{\rdown}[1]{\ensuremath{_{{\color{myred}\downarrow#1}}}}

\makeatletter
\newcommand{\thickhline}{%
    \noalign {\ifnum 0=`}\fi \hrule height 1pt
    \futurelet \reserved@a \@xhline
}
\makeatother
\title{One-Shot Federated Graph Learning via High-Fidelity Proxies and Transferability-Guided Collaboration}

\author{%
  Jia Jiang$^{1}$ \quad
  Zihan Tan$^{1,2}$ \quad
  Wenke Huang$^{3}$ \quad
  Bin Yang$^{1}$ \quad
  Mang Ye$^{1,2}$\thanks{Corresponding author.} \\
  $^{1}$ School of Computer Science, Wuhan University, Wuhan, China. \\
  $^{2}$ Guangdong Laboratory of Artificial Intelligence and Digital Economy, Shenzhen, China. \\
  $^{3}$ College of Computing and Data Science, Nanyang Technological University, Singapore. \\
  \texttt{\{jiangjia,zihantan,yangbin\_cv,yemang\}@whu.edu.cn}\\
}

\begin{document}

\maketitle

\begin{abstract}
One-shot Federated Graph Learning (OFGL) has emerged as a communication-efficient paradigm for federated graph learning under strict bandwidth constraints, compressing collaboration into a single transmission round. Nevertheless, this extreme setting suffers from a fundamental \textbf{Single-Transmission Bottleneck}, which limits both the quality and utility of exchanged knowledge. First, while existing methods attempt to distill private graphs into shareable proxies, their aggressive one-shot compression inherently triggers a fidelity crisis. This compression distorts local graph topology and semantics across clients, leading to \textbf{externalized knowledge underexpression}. Second, although the server aggregates these proxies to update global models, current integration schemes suffer from transferability agnosticism. The server cannot reliably identify cross-client collaboration potential, thus resulting in \textbf{integrated knowledge underutilization}. To tackle these dual challenges, we propose FedHFT, an effective One-shot Personalized Federated Graph Learning (OPFGL) framework that addresses this bottleneck from two perspectives, namely High-Fidelity Knowledge Externalization and Transferability-Ware Knowledge Integration. We use Fidelity-Driven Proxy Refinement to preserve client-side knowledge fidelity, and Transferability-Steered Personalized Collaboration to enable adaptive server-side collaboration. The superiority of FedHFT is validated through extensive experiments on both homophilic and heterophilic graphs. The code is available at https://github.com/Jumpy-Juice/FedHFT.
\end{abstract}

\section{Introduction}
\begin{wrapfigure}{r}{0.40\textwidth}   
\centering   
\includegraphics[width=\linewidth]{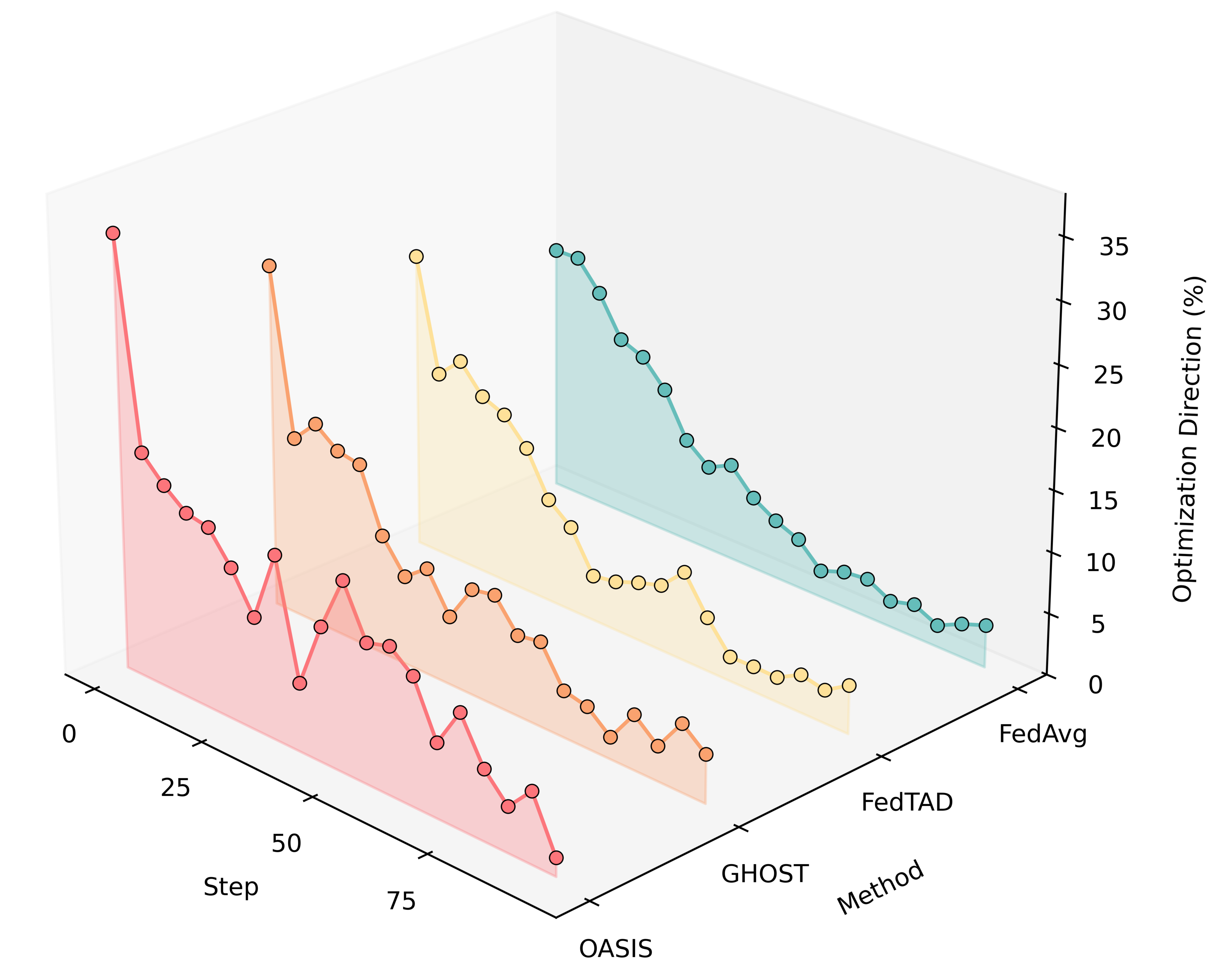}   
\caption{\textbf{Single-Transmission Bottleneck.} With increasing training steps, the optimization gradients induced by existing methods progressively deviate from the true optimization direction, highlighting the bottleneck in the OFGL setting.}  
\label{fig:epoch}  
\end{wrapfigure}

\begin{figure}[t]  
\centering  
\includegraphics[width=\linewidth]{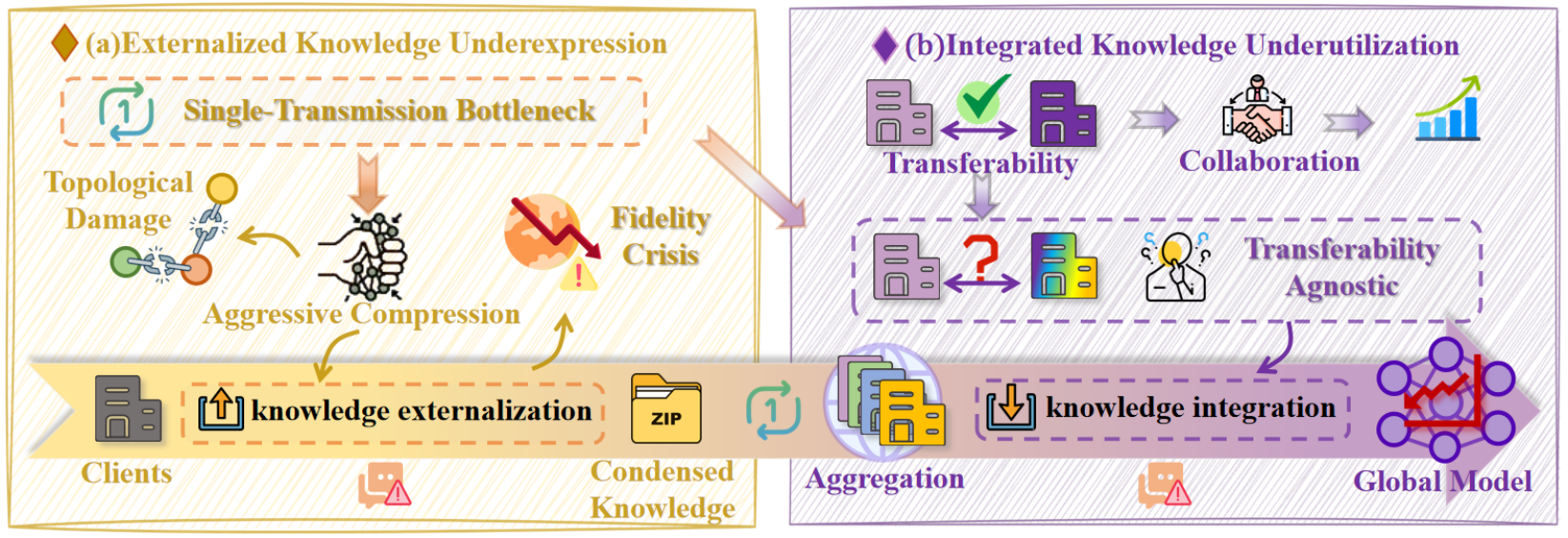}  
\caption{\textbf{Problem Illustration.} The \textbf{Single-Transmission Bottleneck} in OFGL impedes knowledge transfer via two challenges. \textit{(a) Externalized Knowledge Underexpression:} Aggressive compression damages graph topology, causing a \textbf{fidelity crisis}. \textit{(b) Integrated Knowledge Underutilization:} Compressed capacity obscures cross-client collaboration potential, driving \textbf{transferability agnosticism}.} 
\label{fig:problem_illustration}

\end{figure}

Graph-structured data, such as social networks, molecular structures, and knowledge graphs, is ubiquitous and often originates from edge devices \citep{10.3389/fgene.2021.690049,NEURIPS2025_721dbcfe}. The existence of data silos hinders model training, while centralized training poses privacy leakage risks. In this context, Federated Graph Learning (FGL) has emerged \citep{DBLP:journals/sigkdd/FuZDCL22}. It combines the privacy-preserving mechanism of Federated Learning (FL) \citep{pmlr-v54-mcmahan17a,DBLP:conf/cvpr/HuangY022,NEURIPS2025_cdd30bf1,NEURIPS2025_a0f30168,NEURIPS2025_f6913c4f,wan2025energy,shi2025privacy,jia2026consistency} and the powerful graph modeling capabilities of Graph Neural Networks (GNN) \citep{DBLP:journals/isci/0004JWLJJH22}, making it an effective machine learning paradigm \citep{DBLP:journals/pami/HuangYSWLDY24,NEURIPS2025_dcc0ac74,NEURIPS2025_d8ace30c,NEURIPS2025_2d945a80}. Furthermore, FGL is often communication-constrained in practice. To avoid the high communication overhead caused by frequent parameter exchanges and to achieve efficient training under extremely low communication budgets, One-shot Federated Graph Learning (OFGL) has garnered significant attention. However, while OFGL minimizes communication overhead, it involves far less knowledge in the overall training process compared to multi-round interactions \citep{NEURIPS2025_ba0dd552}. It leads to the \textbf{Single-Transmission Bottleneck} as Fig.\ref{fig:epoch} demonstrates. Presented in Fig.\ref{fig:problem_illustration}, we analyze this from two perspectives of knowledge transfer: \textbf{externalized knowledge underexpression}, which occurs when clients distill private graph data into shareable proxies, and \textbf{integrated knowledge underutilization}, which arises when the server aggregates these proxies to update models.

First, one-shot FL (OFL) faces much stricter communication limitations than common FL \citep{DBLP:conf/ijcai/LiuT0SJL0J025,DBLP:conf/ijcai/LiHS21,DBLP:conf/nips/Wang00G024,NEURIPS2025_bb203e93,NEURIPS2025_7ccae68f}. It can only upload highly compressed or distilled proxy knowledge in a single interaction for global collaboration \citep{DBLP:conf/nips/0081C0L0DS022,DBLP:conf/iclr/HeinbaughLS23,DBLP:conf/aistats/Jhunjhunwala0J24}. To meet the data compression requirements under a single communication limit, existing OFL methods primarily rely on strategies such as parameter distillation, feature space alignment, or global statistic integration \citep{DBLP:conf/nips/AllouahDGGKP0024,NEURIPS2025_5df817c5}. While these strategies reduce the communication burden, they cause severe fidelity loss when applied to graph data \citep{DBLP:conf/icml/QianW0ZW0Y25}. Aggressive compression damages inherent topological information and node correlations \citep{DBLP:conf/kdd/TanWHYY26,NEURIPS2025_659afd14}. Specifically, a single transmission has limited information capacity. The compression process inevitably introduces information loss, creating a significant deviation between the proxy knowledge and the original data distribution. We refer to this as the fidelity crisis. Consequently, we propose question 1) How can we achieve \textbf{high-fidelity} in the \textbf{knowledge externalization} phase?

On the other hand, due to the lack of an iterative correction mechanism over multiple rounds, OFGL struggles to recover the complete knowledge structure embedded in client data \citep{DBLP:conf/nips/AllouahDGGKP0024,NEURIPS2025_29978617}. Traditional federated learning employs a global aggregation strategy, assigning uniform weights to all clients for parameter alignment \citep{pmlr-v54-mcmahan17a} . This allows local models to slowly converge towards the common knowledge across clients through multi-round averaging \citep{NEURIPS2025_39540967,NEURIPS2025_29c71a71,NEURIPS2025_7d3f1540,NEURIPS2025_06d2e62a}. However, in OFGL scenarios, a single round eliminates the possibility of iterative refinement. Applying the global aggregation strategy under such constraints not only fails to extract effective common knowledge but also risks catastrophic feature misalignment \citep{DBLP:conf/icml/ZengHZWW0C25,NEURIPS2025_3e8f5d29,NEURIPS2025_aaec5dee,NEURIPS2025_d611d06e,NEURIPS2025_6a9b9b04,NEURIPS2025_0e614c59}. In fact, clients with relevant domain features or complementary information often serve as high-quality collaborators for each other, providing greater positive gains for model updates \citep{DBLP:conf/nips/YangPW00PW0F24,NEURIPS2025_a2582a9f,NEURIPS2025_76ee49ee}. Nevertheless, due to the extremely limited and highly compressed knowledge capacity in one-shot communication, the server struggles to perceive the knowledge transfer potential across client data distributions and build an effective collaboration network when adopting personalized aggregation \citep{DBLP:conf/nips/Ma00ZM24}. This leads to the challenge of transferability agnosticism. Thus, we introduce question 2) How can we accurately assess and utilize client \textbf{transferability}  in the \textbf{knowledge integration} phase?

\newpage
For question 1), in the client-side knowledge externalization phase, we propose a Fidelity-Driven Proxy Refinement (FPR) mechanism. It reduces graph topology and semantic distortion during transmission.  We utilize the Fisher Information Matrix (FIM) as a sensitivity scale \citep{DBLP:conf/cvpr/TanWHLZ0Y25}, aligning graph structural information by introducing a gradient matching constraint. Simultaneously, we select the top-$M$ dimensions most semantically relevant to each class to align global features with local prototypes \citep{DBLP:conf/nips/ZhangSH24}, thereby preserving graph semantic information \citep{10.1145/3539597.3570377}. This mechanism implicitly preserves topology and explicitly purifies semantics, unlike destructive blind compression. It maximizes the information density of the proxy knowledge, protecting the topological information and node correlations in the graph,  maintaining fidelity.

For question 2), in the server-side knowledge integration phase, we propose a Transferability-Steered Personalized Collaboration (TPC) mechanism. It handles data heterogeneity and collaborative matching under one-shot communication. Specifically, the server performs global forward inference on the prototypes uploaded by clients, extracting the negative log-likelihood (NLL) loss vector for each class \citep{DBLP:journals/tmc/YeCHCC26}. By quantifying the distribution differences among these loss vectors, a global similarity matrix is obtained, which drives a dynamic personalized aggregation mechanism. This mechanism accurately assesses client transferability from a global perspective during the extreme single interaction. It implicitly matches each client with high-gain collaborative teammates, ensuring highly personalized delivered models and full utilization of local knowledge.

In summary, the main contributions of this paper are as follows:
\begin{itemize}    
\item We pinpoint two challenges in the Single-Transmission Bottleneck of OFGL: The fidelity crisis of externalized knowledge and transferability agnosticism of integrated knowledge.    
\item We propose an efficient OPFGL framework, FedHFT, maintaining fidelity on the client side and evaluating transferability on the server side.
\item Extensive experimental validation demonstrates that FedHFT significantly outperforms existing baseline methods under one-shot communication.
\end{itemize}

\section{Related Work}
\subsection{Federated Learning}
Federated Learning (FL) enables privacy-preserving distributed collaborative training, yet it confronts the severe challenge of non-independent and identically distributed (non-IID) data \citep{DBLP:conf/nips/0001MO20,NEURIPS2025_e2994aef,DBLP:conf/cvpr/LeeJKOY24,NEURIPS2025_4a4d5fca}. To tackle this problem, existing research enhances the alignment capacity of global models from the dimensions of bottom-level optimization and high-level representation (e.g., FedProx \citep{DBLP:conf/mlsys/LiSZSTS20}, FedNova \citep{DBLP:conf/nips/WangLLJP20}, MOON \citep{DBLP:conf/cvpr/LiHS21}, FedGPS \citep{NEURIPS2025_87ee1bba}, FedEL \citep{NEURIPS2025_3327377e,shi2026don,wu2026swepolyvisionbenchmarkingcrossimageabductive,wu2026sweprometheusmeasuringengineeringgovernance,tan2026metarsirsi2metarecursive}). For highly heterogeneous scenarios, Personalized Federated Learning (pFL) further customizes optimal local models via architecture decoupling (e.g., FedCP \citep{DBLP:conf/kdd/ZhangHWSXMG23}, FedRoD \citep{DBLP:conf/iclr/ChenC22}, APPLE \citep{DBLP:conf/ijcai/0010W22}) or parameter control and state preservation (e.g., FedALA \citep{DBLP:conf/aaai/ZhangHWSXMG23}, FedPHP \citep{DBLP:conf/pkdd/LiZSLS21}, FedWBA \citep{NEURIPS2025_6d0c18c6}). Nevertheless, failing to effectively capture complex graph topological structures and node correlation information, the aforementioned traditional FL and pFL methods generally cannot maintain their superiority in Federated Graph Learning (FGL) scenarios \citep{DBLP:conf/iccv/Feng0LXKZ23, DBLP:conf/iccv/HuTKJ23, DBLP:conf/iccv/SunM0W023, DBLP:conf/cvpr/WangFKWLG24,NEURIPS2025_f2e61505}.

\subsection{Federated Graph Learning}
Federated Graph Learning (FGL) \citep{DBLP:journals/tpds/ChenLMW22} applies the principles of federated learning to graph neural networks \citep{DBLP:conf/cvpr/Hayder024, DBLP:conf/cvpr/LiWGY024, DBLP:journals/nn/LiLLQW24, DBLP:journals/nn/LiWLHL22}. Existing research generally falls into two tracks: inter-graph methods (e.g., GCFL+ \citep{DBLP:conf/nips/XieMXY21}, FedStar \citep{DBLP:conf/aaai/TanLL00Z23}) that boost the generalization of disjoint graphs, and intra-graph methods (e.g., FedSSP \citep{DBLP:conf/nips/TanWHY24,tan2025s2fgl}, FGGP \citep{DBLP:conf/aaai/WanHY24}) that tackle global subgraph collaboration. Besides, targeting the unique attribute challenges of graph data, research such as FedSPA \citep{DBLP:conf/cvpr/TanWHLZ0Y25} further explores the bias issues brought by homophily level heterogeneity \citep{NEURIPS2025_5ee643ae,liu2026fedsdr,shi2026towards,guo2026fedfst,wang2026e2gen}. Although the above methods have been extensively explored in processing graph data \citep{DBLP:conf/cvpr/DuptyDLFGLL24, DBLP:conf/cvpr/Gan0J00L24, DBLP:conf/cvpr/0005SZZ024}, aggressive knowledge compression severely destroys graph structure information when facing extreme communication constraints \citep{shi2025eagles}. Inspired by this, this paper focuses on research from the perspective of maintaining the fidelity of graph knowledge externalization.

\newpage
\subsection{One-shot Federated Graph Learning}
One-shot Federated Graph Learning (OFGL) acts as an efficient solution by reducing communication rounds and enhancing security against eavesdropping attacks. Recent methods \citep{DBLP:conf/nips/0081C0L0DS022, DBLP:conf/iclr/HeinbaughLS23, DBLP:conf/aistats/Jhunjhunwala0J24, DBLP:conf/aaai/YangSLX24} utilize knowledge distillation or neuron-matching techniques to optimize the global model. For graph data, the GHOST framework \citep{DBLP:conf/icml/QianW0ZW0Y25} explores the dilemma of dual heterogeneity across clients. However, existing OFGL methods excessively compress the knowledge uploaded by clients, making the server unable to realize complete expression and full utilization of knowledge. In this study, we adopt an innovative approach that focuses on the transferability among clients, aiming to maximize the collaborative gain of local knowledge in the single aggregation phase.

\section{Preliminary}
\subsection{One-shot Federated Graph Learning and Personalized Aggregation}
In one-shot federated graph learning, a central server coordinates a set of clients $\mathcal{C}$ with $|\mathcal{C}|=K$, indexed by $k$. Each client $k$ owns a private local graph dataset $\mathcal{G}^{k}=(V^{k},E^{k})$ with adjacency matrix $\mathbf{A}^{k}$. For each node $v_i \in V^{k}$, its feature and label are denoted by $\mathbf{x}_i^{k}$ and $y_i^{k}$, respectively. Each client maintains a local differentiable model parameterized by $\theta^{k}$ and optimizes it on private graph data. Unlike conventional iterative FGL, OFGL allows only one communication round: each client externalizes compact local knowledge once and uploads it to the server.

Personalized aggregation at the server aims to deliver a client-specific model for each client \cite{DBLP:conf/ijcai/JiaoZXTZW25,NEURIPS2025_9421261e}.
Given uploaded proxy knowledge from all clients, the server performs one-shot integration and obtains $\Theta^{\text{out}}=\{\theta^{1}_{\text{out}},\ldots,\theta^{K}_{\text{out}}\}$, where $\theta^{k}_{\text{out}}$ is the model dispatched to client $k$. This integration can be written as $\Theta^{\text{out}}=\mathcal{A}(\widetilde{\mathcal{G}})$, where $\widetilde{\mathcal{G}}=\{\widetilde{\mathcal{G}}^{1},\ldots,\widetilde{\mathcal{G}}^{K}\}$ is the set of client proxies and $\mathcal{A}(\cdot)$ is the server-side personalized aggregation operator.
Compared with uniform global aggregation, personalized aggregation explicitly addresses client heterogeneity by constructing tailored model outputs for different clients \cite{NEURIPS2025_66808849,NEURIPS2025_67496dfa}.

\subsection{Proxy Knowledge Generation}
To avoid transmitting raw local graphs in OFGL, each client externalizes compact proxy knowledge through a local conditional generator \cite{DBLP:conf/icml/QianW0ZW0Y25}. For client $k$, Gaussian noise and label conditions are used as inputs, i.e., $\mathbf{Z}^{k} \sim \mathcal{N}(\mathbf{0},\mathbf{I}_{d_z})$ and $\mathbf{Y}^{k}$, and the generator $g_{\phi^{k}}$ produces synthetic node representations:
\[
\widetilde{\mathbf{X}}^{k} = g_{\phi^{k}}(\mathbf{Z}^{k},\mathbf{Y}^{k}).
\]

The generator is optimized locally so that the generated representations preserve task-relevant information from private data while remaining communication-efficient.
Based on $\widetilde{\mathbf{X}}^{k}$, the client constructs proxy edges from feature-space relations and obtains a local proxy graph
\[
\widetilde{\mathcal{G}}^{k}=(\widetilde{\mathbf{X}}^{k},\widetilde{\mathbf{E}}^{k}),
\]
which is used for local alignment training with the private graph.
After local training, each client uploads only compact proxy knowledge rather than raw nodes, raw edges, or full training trajectories.
Upon receiving proxies from all clients, the server reconstructs corresponding pseudo graphs from the uploaded knowledge and performs one-shot model integration on these pseudo graphs.
Therefore, proxy knowledge generation serves as the bridge between local private optimization and server-side collaborative learning under a strict single-round communication budget.

\section{Methodology}
\paragraph{Overview.}
One-shot federated graph learning faces a Single-Transmission Bottleneck, causing a fidelity crisis in knowledge externalization and transferability agnosticism in knowledge integration. Our coupled framework addresses these vulnerabilities. Locally, FPR enhances the fidelity of uploaded proxy knowledge via structural and semantic alignment. Globally, TPC resolves agnosticism by evaluating inter-client transferability from the uploads to drive personalized aggregation. Each client uploads its generator weights and class-level proxy prototypes once (overhead and privacy analyses are in Appendix~\ref{overhead} and~\ref{privacy}).

\subsection{Client-side FPR: Fidelity-Driven Proxy Refinement}
\begin{figure*}[t]
  \centering
  \includegraphics[width=\textwidth]{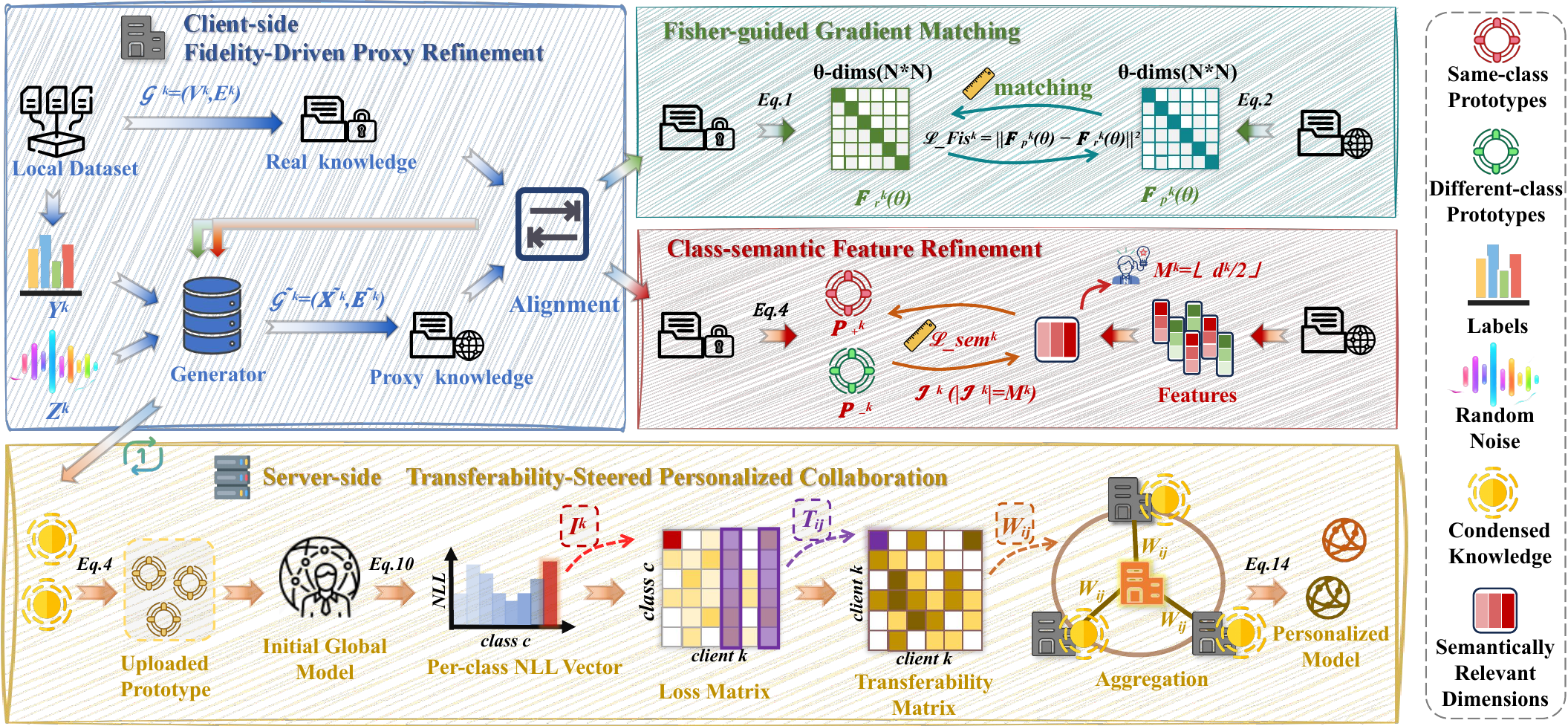}
  \caption{\textbf{Architecture of FedHFT} for one-shot federated graph learning. \textit{Top: Client-side FPR.} Clients generate high-fidelity proxy knowledge via Fisher-guided gradient matching and top-$M$ class-semantic feature refinement prior to the single upload. \textit{Bottom: Server-side TPC.} The server performs personalized aggregation using inter-client transferability matrices derived from per-class NLL of the uploaded prototypes.}
  \label{fig:framework}
  
\end{figure*}

\paragraph{Motivation.}
Existing proxy generators fail to align parameter-gradient responses at the optimization-trajectory level between proxy and real training. Furthermore, full-dimensional semantic constraints inevitably mix class-critical signals with noisy dimensions. In one-shot communication, such local mismatch is irreparable on the server. Therefore, FPR introduces Fisher-guided matching to align optimization trajectories, alongside top-$M$ class-semantic alignment focused strictly on gradient-selected informative dimensions, filtering out noisy channel interference before upload. Details are presented in Fig.\ref{fig:framework}.

\subsubsection{Fisher-guided Gradient Matching}
The key idea of this module is to reduce the optimization mismatch between real local training and proxy-based training.
In OFGL, the server can only observe uploaded proxy knowledge, so if the proxy path induces a different parameter-update tendency from the real-data path, this bias will be directly transferred to the server and cannot be corrected later.

To make this mismatch measurable, we construct two Fisher statistics for each client $k$ and parameter $\theta$ over a batch of size $B$.
$\mathbf{F}_{r}^{k}(\theta)$ is computed from real local data and reflects how strongly parameter $\theta$ responds to real supervision:
\begin{equation}
\mathbf{F}_{r}^{k}(\theta)=\left(\frac{1}{B}\sum_{b=1}^{B}\frac{\partial \log p_{\theta}(y_b^{k}\mid \mathbf{x}_b^{k})}{\partial \theta}\right)^2.
\end{equation}
In implementation, this corresponds to accumulating log-likelihood gradients on local real data and squaring averaged gradients. In parallel, we compute $\mathbf{F}_{p}^{k}(\theta)$ on the proxy path:
\begin{equation}
\mathbf{F}_{p}^{k}(\theta)=\left(\frac{1}{B_p}\sum_{b=1}^{B_p}\frac{\partial \log p_{\theta}(\tilde y_b^{k}\mid \tilde{\mathbf{x}}_b^{k})}{\partial \theta}\right)^2.
\end{equation}
This branch is evaluated by retaining the computation graph to compute higher-order derivatives, so the discrepancy term can backpropagate through the generation process and directly optimize the generator. The alignment objective is defined as
\begin{equation}
\mathcal{L}_{\text{Fis}}^{k}
=\sum_{\theta\in\Theta}\left(\mathbf{F}_{p}^{k}(\theta)-\mathbf{F}_{r}^{k}(\theta)\right)^2.
\end{equation}
Here, $\Theta$ denotes the model parameter set used in Fisher matching.
Minimizing this term encourages proxy-induced parameter responses to follow real-data-induced responses, thereby reducing trajectory-level deviation before upload. The comprehensive explanation of Fisher matching is provided in Appendix~\ref{Fisher_matching}.

\subsubsection{Class-semantic Feature Refinement}
The goal of this module is to enforce semantic alignment only on informative channels, rather than on the full feature space.
In graph data, class-discriminative semantics and structure-dependent signals are entangled. Forcing all dimensions to match class prototypes can mix useful class cues with noisy channels, weaken anomalous-node expression, and interfere with normal nodes that rely on neighborhood-dependent information.
Therefore, we first estimate topology-aware class priors, then select gradient-informative channels, and finally apply class-semantic alignment only on the selected top-$M$ dimensions.

For client $k$, topology importance is computed by PageRank:
\begin{equation}
\label{eq:pagerank}
\mathbf{v}^{k,(t+1)}=\alpha \mathbf{T}^{k}\mathbf{v}^{k,(t)}+\frac{1-\alpha}{N^{k}}\mathbf{1}.
\end{equation}
Where $\mathbf{T}^{k}$ is the local transition matrix, $N^{k}$ is the local node number, and $\alpha=0.85$.
The term $\alpha \mathbf{T}^{k}\mathbf{v}^{k,(t)}$ propagates node importance through local connectivity, while $\frac{1-\alpha}{N^{k}}\mathbf{1}$ is a restart term that stabilizes the computation.
After a fixed number of local power-iteration steps, we obtain topology-aware node importance scores for client $k$.
These scores are then used to build class priors from \emph{real} local features (not generated features):
\begin{equation}
\mathbf{p}_{c}^{k}
=\frac{\sum_{i\in\mathcal{V}_{c}^{k}} v_{i}^{k}\mathbf{x}_{i}^{k}}
{\sum_{i\in\mathcal{V}_{c}^{k}} v_{i}^{k}+\epsilon}.
\end{equation}
Here, $\mathcal{V}_{c}^{k}$ denotes nodes of class $c$ on client $k$, and each node feature $\mathbf{x}_{i}^{k}$ contributes proportionally to its topology weight $v_{i}^{k}$. $\epsilon$ is a small constant for numerical stability.
This weighted prototype acts as a class prior that emphasizes structurally central evidence.
Compared with plain averaging, it suppresses contributions from weakly connected or less informative nodes and yields a more topology-consistent class summary for subsequent semantic alignment.
With topology-aware class priors, we identify informative dimensions from the generated features by gradient probing. Specifically, we detach generated node logits, enable gradient tracking, and compute a probe loss $\mathcal{L}_{\text{probe}}$ (cross-entropy on generated nodes) to obtain gradients with respect to generated features.
Channel salience is defined as
\begin{equation}
\mathbf{g}^{k}=\operatorname{mean}_{i}\left|\frac{\partial \mathcal{L}_{\text{probe}}}{\partial \widetilde{\mathbf{x}}_{i}^{k}}\right|.
\end{equation}
For each feature channel, this quantity measures how strongly the probe classification loss reacts when that channel changes.
A larger value indicates that this channel is more influential to the local decision boundary and thus should be prioritized in semantic refinement.

Then we select the Top-$M$ channels, setting $M = \lfloor \frac{d^k}{2} \rfloor$, where $d^k$ is the feature dimension of client $k$. This fixed refinement budget focuses alignment on high-impact semantic dimensions instead of spreading constraints over the full space.
This selective strategy is crucial for avoiding noisy-channel interference in graph scenarios where semantic and structural factors are mixed.
On the selected channels, we compute class-semantic positive/negative KL terms. A sensitivity analysis of Top-$M$ can be found in Appendix~\ref{topm}.

For each generated sample on client $k$, we construct class-semantic positive and negative prototype matrices by indexing class prototypes according to its label and a mismatched label, respectively. We denote them as $\mathbf{P}_{+}^{k}$ and $\mathbf{P}_{-}^{k}$, where $\mathbf{P}_{+}^{k}$ contains same-class prototypes and $\mathbf{P}_{-}^{k}$ contains different-class prototypes aligned sample-wise with $\widetilde{\mathbf{X}}^{k}$.
On the selected channel index set $\mathcal{I}^{k}$ with $|\mathcal{I}^{k}|=M^k$, we define
\[
q^{k}=\operatorname{softmax}\!\big(\widetilde{\mathbf{X}}^{k}[:,\mathcal{I}^{k}]\big),\quad
p_{+}^{k}=\operatorname{softmax}\!\big(\mathbf{P}_{+}^{k}[:,\mathcal{I}^{k}]\big),\quad
p_{-}^{k}=\operatorname{softmax}\!\big(\mathbf{P}_{-}^{k}[:,\mathcal{I}^{k}]\big).
\]
\begin{equation}
\mathcal{L}_{\text{pos}}^{k}=D_{\mathrm{KL}}(q^{k}\|p_{+}^{k}),\qquad
\mathcal{L}_{\text{neg}}^{k}=D_{\mathrm{KL}}(q^{k}\|p_{-}^{k}).
\end{equation}
Here, all three distributions are constructed on the selected feature channels (i.e., the top-$M$ dimensions), where $p_{+}^{k}$ is the same-class prototype distribution and $p_{-}^{k}$ is a different-class prototype distribution.
We therefore impose a contrastive semantic constraint: generated features should remain close to their class-consistent prior while being separated from mismatched classes.
The resulting top-$M$ class-semantic alignment term is
\begin{equation}
\mathcal{L}_{\text{sem}}^{k}
=\lambda_{\text{sem}}\exp\!\left(\mathcal{L}_{\text{pos}}^{k}-\mathcal{L}_{\text{neg}}^{k}\right).
\end{equation}
Where \(\lambda_{\text{sem}}\) is a trade-off hyperparameter controlling the strength of class-semantic alignment.

Minimizing this term encourages generated representations to stay closer to the same-class priors than the different-class priors on informative dimensions, reducing semantic drift while avoiding full-dimensional over-constraint.

\subsection{Server-side TPC: Transferability-Steered Personalized Collaboration}
\paragraph{Motivation.}
In OFGL, lack of multi-round correction makes uniform aggregation unreliable, risking severe feature misalignment \cite{NEURIPS2025_16e71d1a,NEURIPS2025_8c4c80d3,NEURIPS2025_cfcb1905}. While some clients offer high collaborative gain to specific peers, highly compressed uploads in one-shot communication obscure these relationships, causing transferability agnosticism in personalized aggregation. Consequently, the core requirement at this stage is to accurately estimate inter-client transferability from the limited uploaded proxy knowledge to construct an effective personalized collaboration network. Details are presented in Fig.\ref{fig:framework}.

Each client uploads class-level prototypes of proxy knowledge to the server.
At this stage, prototype construction also uses PageRank weighting, and the PageRank update is the same as Eq.\ref{eq:pagerank}; therefore, we directly reuse the same topology-aware formulation.
For client $k$, the uploaded prototype matrix is
\begin{equation}
\mathbf{P}^{k}=[\mathbf{p}_{1}^{k},\ldots,\mathbf{p}_{C}^{k}]^\top ,
\end{equation}
where $C$ is the total number of classes.
These prototypes provide compact class summaries that retain topology-aware semantic cues while avoiding raw graph exposure.
Receiving $\mathbf{P}^{k}$, the server performs global forward inference using the initial global model $f_{\theta}$ and computes class-wise NLL values:
\begin{equation}
\mathbf{l}^{k}[c]=\operatorname{NLL}\!\left(f_{\theta}(\mathbf{p}_{c}^{k}),\,c\right).
\end{equation}
 Here, \(f_{\theta}\) is obtained by determining the model architecture based on task configuration and dataset statistics, and initializing its parameters under a fixed random seed. Stacking all client vectors yields
\begin{equation}
\mathbf{L}=
\begin{bmatrix}
(\mathbf{l}^{1})^\top\\
(\mathbf{l}^{2})^\top\\
\vdots\\
(\mathbf{l}^{K})^\top
\end{bmatrix}\in\mathbb{R}^{K\times C}.
\end{equation}
This stacked loss matrix $\mathbf{L}$ serves as a behavior-level representation of uploaded knowledge under a common model view. In one-shot heterogeneous settings, minority-class prototypes derived from scarce samples lack multi-round refinement and often degenerate into unreliable near-random vectors, rendering direct geometric metrics (e.g., MMD \cite{NEURIPS2025_74fc5575} or Cosine Similarity) highly misleading. To circumvent this spatial fragility, we evaluate class-wise NLL via the globally initialized model $f_\theta$. This mapping translates noisy prototypes into output logits, acting as an advanced representation of the label distribution. By assessing class-wise distributional strength rather than unstable spatial coordinates, this loss-based mechanism provides a highly robust proxy for inter-client transferability.

Based on $\mathbf{L}$, transferability between clients $i$ and $j$ is defined as
\begin{equation}
T_{ij}
=\cos\!\big(\mathbf{l}^{i,(\cap)},\mathbf{l}^{j,(\cap)}\big)\cdot
\frac{|\mathcal{Y}^i\cap\mathcal{Y}^j|}{|\mathcal{Y}^i\cup\mathcal{Y}^j|}.
\end{equation}
Here, \(\mathcal{Y}^i\) denotes the set of valid classes on client \(i\), and \(\mathbf{l}^{i,(\cap)}\) and \(\mathbf{l}^{j,(\cap)}\) denote loss vectors restricted to the overlapping class set \(\mathcal{Y}^i \cap \mathcal{Y}^j\).
The cosine component measures whether two clients exhibit similar class-wise loss patterns, while the Jaccard term downweights pairs with weak class overlap.
This combination prevents unreliable relations from being overestimated and produces a transferability matrix \(\mathbf{T}\in\mathbb{R}^{K\times K}\) that is both behavior-aware and label-overlap-aware.
To convert transferability into actionable collaboration strength, we normalize each row by temperature-scaled softmax:
\begin{equation}
W_{ij}=\frac{\exp(T_{ij}/\tau)}{\sum_{j'}\exp(T_{ij'}/\tau)}.
\end{equation}
Where $\tau$ is a temperature hyperparameter.
The resulting weight matrix \(\mathbf{W}\) makes each target client emphasize highly transferable sources and suppress weakly related ones. The server synthesizes pseudo-graphs using the complete set of uploaded generator parameters, and subsequently performs personalized integration through transferability-driven weighted supervision:
\begin{equation}
\mathcal{L}_{i}=\sum_{j}\mathcal{L}_{i\leftarrow j}
=\sum_{j}W_{ij}\,\mathcal{L}_{\text{sup}}^{(i\leftarrow j)}.
\end{equation}
Here, $\mathcal{L}_{\text{sup}}^{(i\leftarrow j)}$ denotes the supervised loss of the target personalized model for client $i$ evaluated on the uploaded proxy graph from client $j$.
After optimization, the server dispatches personalized models to corresponding clients.
Through this process, TPC uses limited one-shot uploaded knowledge to assess client transferability from a global perspective. It builds a personalized collaboration network, identifies positive-gain collaborators for each client, and aggregates models that better match each client’s local data characteristics. 

\section{Experiments}
\subsection{Experimental Setup}

\paragraph{Datasets.}
We evaluate our method on seven benchmark graph datasets for node classification tasks: Cora \cite{DBLP:journals/ir/McCallumNRS00}, CiteSeer \cite{DBLP:conf/dl/GilesBL98}, PubMed \cite{canese2013pubmed}, Actor \cite{DBLP:conf/iclr/PeiWCLY20}, Amazon-Photo \cite{DBLP:journals/corr/abs-1811-05868}, Roman-empire \cite{DBLP:conf/iclr/PlatonovKDBP23}, and Ogbn-Arxiv \cite{DBLP:conf/nips/HuFZDRLCL20}. Specifically, Cora, CiteSeer, and PubMed are citation networks; Amazon-Photo is an Amazon co-purchase graph; Actor and Roman-empire are commonly used heterophilous benchmarks; and Ogbn-Arxiv is a large-scale graph benchmark. These datasets jointly cover diverse graph scales and structural properties for a comprehensive evaluation. Detailed descriptions and splits for these datasets can be found in Appendix~\ref{Dataset_details}.

\paragraph{Baselines.}
We compare our method against 14 representative baselines: (1) FedAvg~\cite{pmlr-v54-mcmahan17a} as the canonical FL algorithm; (2) FedProx~\cite{DBLP:conf/mlsys/LiSZSTS20} and (3) FedNova~\cite{DBLP:conf/nips/WangLLJP20} as robust optimization and personalization methods under data heterogeneity; (4) FedPub~\cite{DBLP:conf/icml/BaekJJYH23}, (5) FedTAD~\cite{DBLP:conf/ijcai/ZhuLWWHL24}, (6) FedGTA~\cite{DBLP:journals/pvldb/LiWZZLW23}, (7) FGGP~\cite{DBLP:conf/aaai/WanHY24}, and (8) FGSSL~\cite{DBLP:conf/ijcai/HuangWY023} as recent state-of-the-art FGL approaches; (9) DENSE~\cite{DBLP:conf/nips/0081C0L0DS022}, (10) FedCVAE~\cite{DBLP:conf/iclr/HeinbaughLS23}, (11) FedSD2C~\cite{NEURIPS2024_ba0ad9d1}, and (12) FedFisher~\cite{DBLP:conf/aistats/Jhunjhunwala0J24} as data-free and one-shot FL baselines; and (13) GHOST~\cite{DBLP:conf/icml/QianW0ZW0Y25} and (14) OASIS~\cite{wan2026oasis} as highly relevant OFGL competitors. Detailed descriptions of all baselines can be found in the Appendix~\ref{baseline}.

\paragraph{Implementation Details.}
We use a two-layer GCN backbone with hidden size 128 and dropout 0.3. 
The federated setting uses \(K=10\) clients, where client label proportions are sampled from a Dirichlet distribution with \(\alpha=0.05\) to simulate non-IID data. 
Each client performs local optimization with Adam for \(T_L=100\) steps, and experiments are conducted with one communication round. For performance evaluation, we record the average local accuracy across all clients to comprehensively reflect the personalized local performance. To ensure statistical reliability, all reported results are averaged over at least ten independent runs with different random seeds.
More implementation details and additional studies are provided in the Appendix~\ref{implementation}.

\subsection{Experimental Results}
\paragraph{Performance Comparison}Across diverse non-IID federated graph learning settings , FedHFT consistently shows better overall generalization and stability than existing baselines, indicating stronger robustness to distribution shifts. As Tab.\ref{tab:performance} demonstrates, while OFGL methods are generally more competitive than traditional FL, FGL and OFL approaches, they still remain inferior to FedHFT overall. Notably, FedHFT exhibits exceptional effectiveness on large-scale and complex graphs compare to recent state-of-the-art one-shot methods. Additional experimental results under different settings are provided in the Appendix~\ref{different number} and Appendix~\ref{dir}. A comparison under matched total communication budgets is further reported in Appendix~\ref{equal_budget}.

\begin{table*}[t]

\caption{Comparison with state-of-the-art baselines on seven datasets. We report node classification accuracy (\%) for downstream performance evaluation. Green arrows \gup{} indicate improvements over FedAvg, while red arrows \rdown{} denote performance drops relative to FedAvg. OOM means out-of-memory error. The \textbf{best} and \underline{second-best} results in each column are marked in bold and underline, respectively. Among the baselines, FedPub and FedGTA are personalized methods, while the others learn a single global model.}

\centering
\renewcommand\arraystretch{1.5}
\resizebox{\textwidth}{!}{
\begin{tabular}{c|l||ccccccc} 
\hline\thickhline
\rowcolor{mygray}
\textbf{Category} & \textbf{Methods} & \textbf{Cora} & \textbf{CiteSeer} & \textbf{PubMed} & \textbf{Actor} & \textbf{Amz-Photo} & \textbf{Roman-Empire} & \textbf{Ogbn-Arxiv} \\
\hline\hline
\multirow{3}{*}{FL} 
& FedAvg [ASTAT17] & 34.09 & 43.00 & 55.45 & 24.00 & 29.87 & 21.41 & 23.83 \\
\cline{2-9}
& FedProx [MLSys20] & 33.31\rdown{0.78} & 41.84\rdown{1.16} & 43.04\rdown{12.40} & 22.69\rdown{1.30} & 27.44\rdown{2.42} & 19.32\rdown{2.09} & 23.58\rdown{0.25} \\
& FedNova [NeurIPS20] & 15.50\rdown{18.59} & 16.45\rdown{26.55} & 35.33\rdown{20.12} & 20.01\rdown{3.98} & 13.00\rdown{16.86} & 6.64\rdown{14.77} & 3.20\rdown{20.64} \\
\hline
\multirow{5}{*}{FGL} 
& FedPub [ICML23] & 31.06\rdown{3.03} & 24.63\rdown{18.37} & 41.29\rdown{14.15} & 24.19\gup{0.19} & 25.69\rdown{4.18} & 17.06\rdown{4.35} & 14.34\rdown{9.49} \\
& FedTAD [IJCAI24] & 34.62\gup{0.53} & \underline{46.37}\gup{3.36} & 47.40\rdown{8.04} & 21.71\rdown{2.28} & 28.67\rdown{1.20} & 20.34\rdown{1.07} & 11.50\rdown{12.34} \\
& FedGTA [VLDB24] & 15.94\rdown{18.14} & 14.15\rdown{28.86} & 8.27\rdown{47.18} & 14.20\rdown{9.79} & 9.33\rdown{20.53} & 13.29\rdown{8.12} & 8.01\rdown{15.82} \\
& FGGP [AAAI24] & 32.50\rdown{1.59} & 36.85\rdown{6.15} & 49.13\rdown{6.31} & 24.37\gup{0.38} & 26.31\rdown{3.56} & 21.29\rdown{0.12} & OOM \\
& FGSSL [IJCAI23] & 13.85\rdown{20.24} & 19.04\rdown{23.96} & 31.57\rdown{23.88} & 19.53\rdown{4.46} & 10.18\rdown{19.69} & 6.20\rdown{15.21} & 3.02\rdown{20.81} \\
\hline
\multirow{4}{*}{OFL} 
& DENSE [NeurIPS22] & 26.27\rdown{7.82} & 26.91\rdown{16.09} & 46.98\rdown{8.47} & 20.16\rdown{3.83} & 20.22\rdown{9.65} & 15.86\rdown{5.55} & 20.80\rdown{3.04} \\
& FedCVAE [ICLR23] & 31.91\rdown{2.17} & 26.91\rdown{16.10} & 47.67\rdown{7.77} & 19.92\rdown{4.08} & 27.87\rdown{2.00} & 14.93\rdown{6.48} & 11.95\rdown{11.88} \\
& FedSD2C [NeurIPS24] & 28.84\rdown{5.24} & 30.28\rdown{12.72} & 37.58\rdown{17.86} & 19.80\rdown{4.20} & 22.38\rdown{7.49} & 22.62\gup{1.21} & 28.12\gup{4.28} \\
& FedFisher [AISTATS24] & 36.14\gup{2.05} & 34.33\rdown{8.67} & 50.13\rdown{5.32} & 18.88\rdown{5.12} & 34.83\gup{4.96} & 15.74\rdown{5.67} & 4.41\rdown{19.42} \\
\hline 
& GHOST [ICML25] & 33.55\rdown{0.54} & 35.45\rdown{7.55} & 46.35\rdown{9.09} & 23.41\rdown{0.59} & 29.85\rdown{0.02} & 19.90\rdown{1.51} & 14.03\rdown{9.80} \\
& OASIS [NeurIPS25] & \underline{40.05}\gup{5.96} & 42.82\rdown{0.18} & \textbf{65.96}\gup{10.51} & \underline{25.51}\gup{1.51} & \underline{46.04}\gup{16.17} & \underline{23.68}\gup{2.27} & \underline{28.87}\gup{5.04} \\
\cline{2-9}
\rowcolor[HTML]{FFF0C1}
\cellcolor{white}\multirow{-3}{*}{OFGL} & \textbf{FedHFT (ours)} & \textbf{52.24}\gup{18.15} & \textbf{56.02}\gup{13.02} & \underline{56.98}\gup{1.53} & \textbf{41.51}\gup{17.51} & \textbf{49.30}\gup{19.43} & \textbf{40.82}\gup{19.41} & \textbf{44.50}\gup{20.67}  \\
\hline\hline
\end{tabular}
}
\label{tab:performance}

\end{table*}

\subsection{Ablation Study}
\paragraph{Effects of Key Components Mechanism of FedHFT}
To evaluate specific design components, we conduct an ablation study on our two main modules: FPR and TPC. As shown in table~\ref{tab:ablation_fpr_tpc}, employing either FPR or TPC independently significantly enhances accuracy across all datasets. Combining both achieves the optimal performance, validating that high-fidelity externalization (FPR) and transferability-steered integration (TPC) mutually reinforce each other to overcome the single-transmission bottleneck.

\paragraph{Effects of Key Components Mechanism of FPR}
To evaluate the FPR mechanism, we ablate Fisher-guided gradient matching ($\mathcal{L}_{\text{Fis}}$) and class-semantic alignment ($\mathcal{L}_{\text{sem}}$). As shown in table~\ref{tab:ablation_lambda}, while applying either trajectory-level matching or semantic refinement alone brings noticeable improvements, their combination yields optimal gains. This validates FPR's effectiveness in enhancing the fidelity of uploaded proxy knowledge by aligning graph structures and refining semantic proximity.

\begin{table}[!ht]
    \centering
    
    \small
    \begin{minipage}[t]{0.48\linewidth}
        \centering
        \caption{\small\textbf{Ablation study} of FPR and TPC across datasets.}
        \label{tab:ablation_fpr_tpc}
        \resizebox{\columnwidth}{!}{%
            \setlength\tabcolsep{5pt}
            \renewcommand\arraystretch{1.2}
            \begin{tabular}{c|c||cccc}
                \hline \thickhline
                \rowcolor{mygray}
                && \multicolumn{4}{c}{Dataset}\\
                \cline{3-6}
                \rowcolor{mygray}
                \multirow{-2}{*}{FPR} & \multirow{-2}{*}{TPC}
                & cora & citeseer & actor & roman\_empire \\
                \hline\hline
                \ding{55} & \ding{55} & 29.01 & 27.26 & 18.20 & 23.93 \\
                \ding{51} & \ding{55} & 41.35 & 39.46 & 25.36 & 30.11 \\
                \ding{55} & \ding{51} & 45.25 & 54.19 & 34.34 & 34.21 \\
                \rowcolor[HTML]{D7F6FF}
                \ding{51} & \ding{51} & \textbf{52.24} & \textbf{56.02} & \textbf{41.51} & \textbf{40.82} \\
                \hline
            \end{tabular}%
        }
    \end{minipage}%
    \hfill %
    \begin{minipage}[t]{0.48\linewidth}
        \centering
        \caption{\small\textbf{Ablation study} of Fisher matching ($\mathcal{L}_{\text{Fis}}$) and class-semantic alignment ($\mathcal{L}_{\text{sem}}$).}
        \label{tab:ablation_lambda}
        \resizebox{\columnwidth}{!}{%
            \setlength\tabcolsep{5pt}
            \renewcommand\arraystretch{1.2}
            \begin{tabular}{c|c||cccc}
                \hline \thickhline
                \rowcolor{mygray}
                && \multicolumn{4}{c}{Dataset}\\
                \cline{3-6}
                \rowcolor{mygray}
                \multirow{-2}{*}{$\mathcal{L}_{\text{Fis}}$} & \multirow{-2}{*}{$\mathcal{L}_{\text{sem}}$}
                & cora & citeseer & actor & roman\_empire \\
                \hline\hline
                \ding{55} & \ding{55} & 29.01 & 27.26 & 18.20 & 23.93 \\
                \ding{51} & \ding{55} & 37.01 & 36.46 & 24.98 & 29.27 \\
                \ding{55} & \ding{51} & 39.95 & 35.13 & 23.59 & 26.37 \\
                \rowcolor[HTML]{D7F6FF}
                \ding{51} & \ding{51} & \textbf{41.35} & \textbf{39.46} & \textbf{25.36} & \textbf{30.11} \\
                \hline
            \end{tabular}%
        }
    \end{minipage}
    
\end{table}

\subsection{Hyper-parameter Study}
\begin{figure}[!ht]
    \centering
    \begin{subfigure}[b]{0.64\textwidth}
        \centering
        \includegraphics[width=\textwidth]{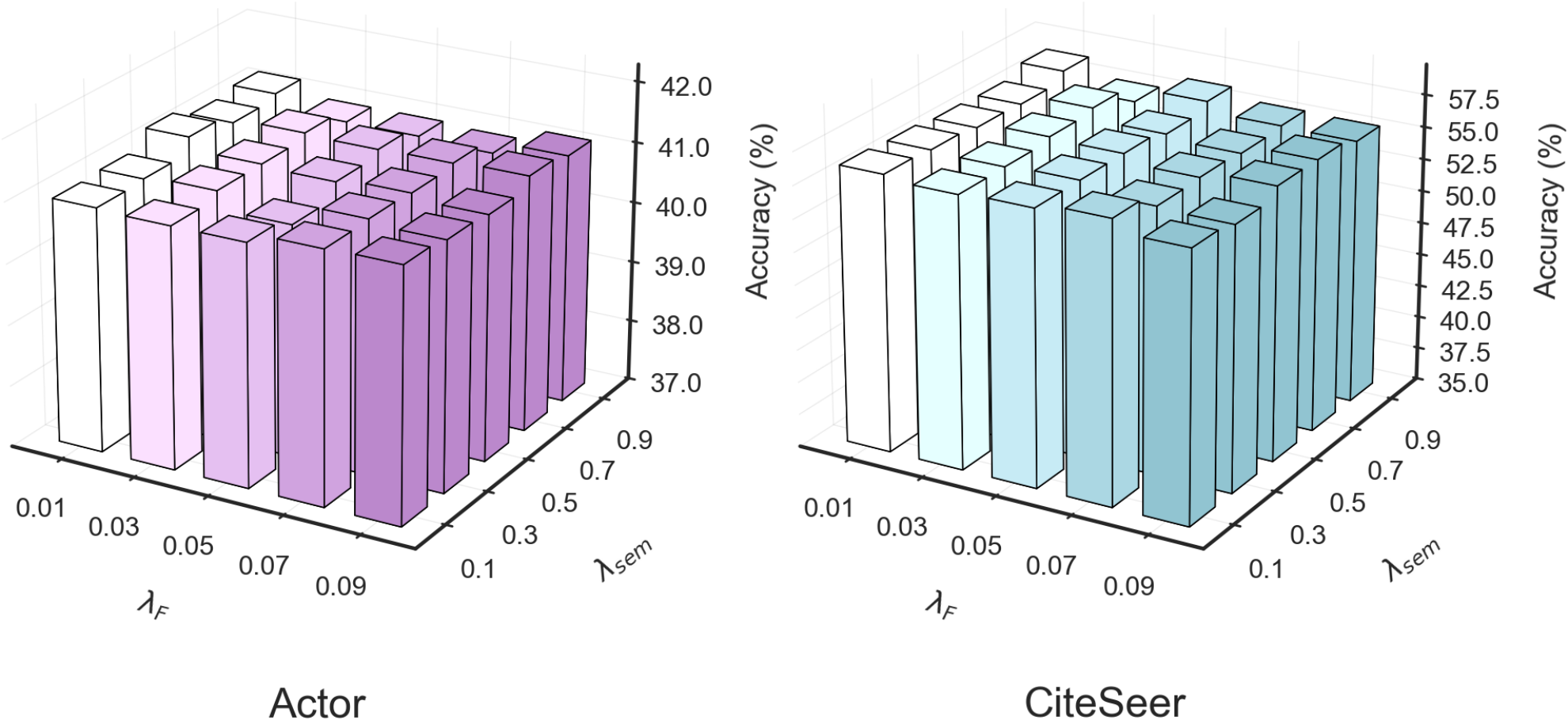}
        \caption{Analysis of $\lambda$}
        \label{fig:sweep}
    \end{subfigure}
    \hfill
    \begin{subfigure}[b]{0.32\textwidth}
        \centering
        \includegraphics[width=\textwidth]{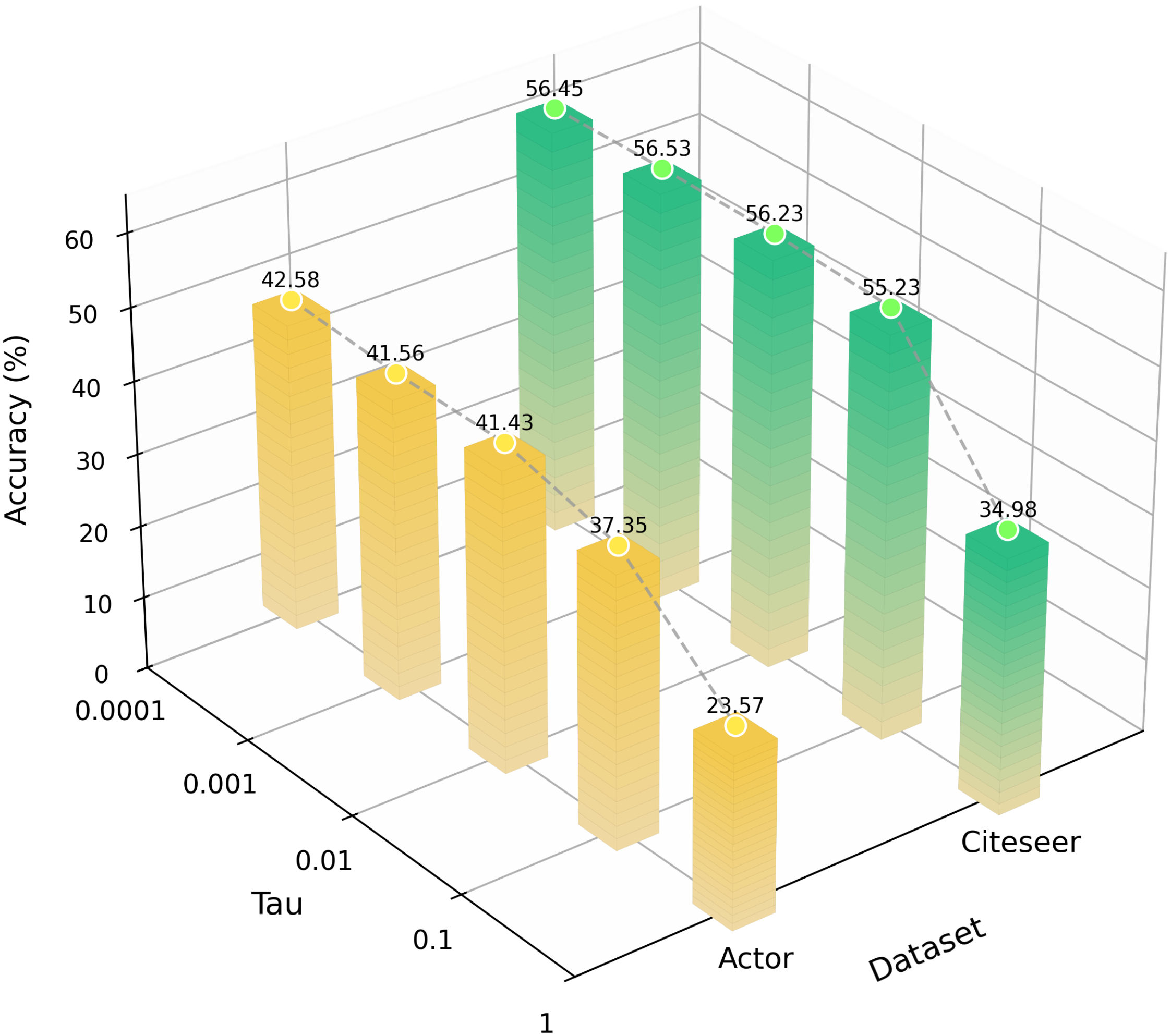}
        \caption{Analysis of $\tau$}
        \label{fig:tau}
    \end{subfigure}
    \caption{\textbf{Analysis on hyper-parameters in FedHFT.} Node classification results under different refinement weights (a) and temperature coefficient (b). The results illustrate the performance sensitivity across representative datasets, demonstrating that FedHFT maintains stable accuracy over a broad range of parameter values. The dashed lines represent the baseline test accuracy for each setting.}
    \label{fig:hyperparameters}
    
\end{figure}

We further conduct a hyper-parameter sensitivity analysis for FedHFT in Fig.\ref{fig:hyperparameters}, investigating the impact of the temperature coefficient $\tau$ from TPC, Fisher matching weight $\lambda_F$, and class-semantic alignment weight $\lambda_{\text{sem}}$. With other configurations fixed, a consistent trend is observed across various datasets and non-IID scenarios: FedHFT achieves stable performance across a broad range of parameter values, but experiences significant degradation at extreme values. This indicates that while our method exhibits strong hyper-parameter robustness, it still relies on appropriate parameter selection. Based on this analysis, our main experiments utilize the optimal hyper-parameters identified on the validation set for each dataset, thereby ensuring fair comparisons and fully demonstrating the model's capabilities. 
\clearpage
\subsection{Superiority Analysis}

\begin{wrapfigure}{r}{0.35\textwidth} 
  \centering
   \vspace{-35pt}
  \includegraphics[width=\linewidth]{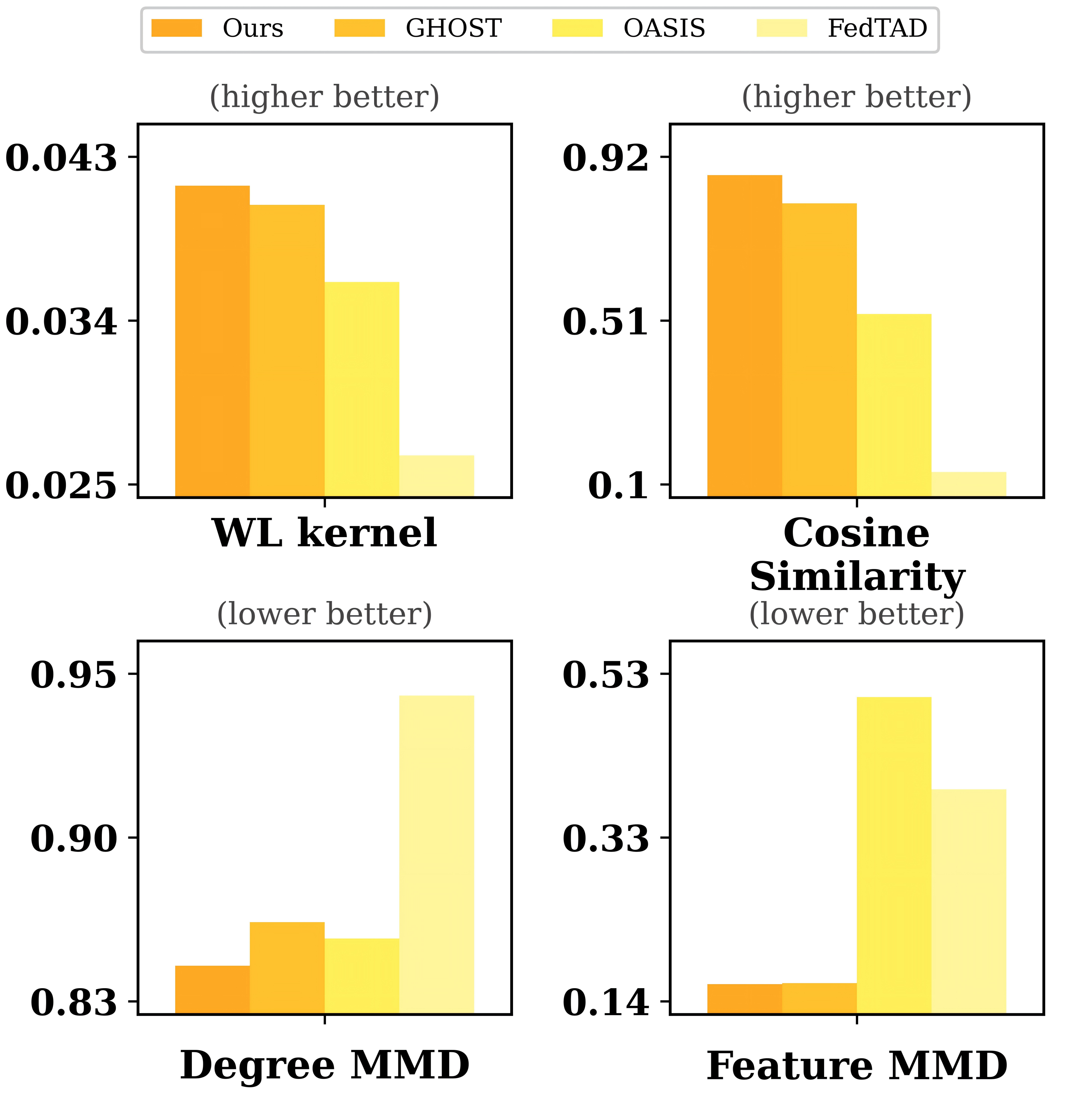}
 
  \caption{\textbf{Proxy Data Analysis.} Our method consistently performs best across all four proxy-quality metrics.}
  \label{fig:proxy}
  \vspace{-10pt}
  
\end{wrapfigure}

To demonstrate the superiority of our method in proxy knowledge generation, we conduct comparisons under a unified experimental setting against GHOST, OASIS, FedTAD, and other methods that also rely on proxy data, and perform a comprehensive analysis from the perspectives of structural preservation and feature preservation. Specifically, the WL kernel characterizes structural similarity between the generated results and the original graph, Cosine Similarity measures the consistency of feature representation directions, Degree MMD reflects differences in node degree distributions, and Feature MMD quantifies shifts in feature distributions. The results in Fig.\ref{fig:proxy} show that our method achieves the best overall performance across these metrics, indicating that it preserves the structural information and feature semantics of the original graph more effectively. The transferability--accuracy correlation is reported in Appendix~\ref{transfer}.

\section{Conclusion}
 In this paper, we propose \textbf{FedHFT} to address the Single-Transmission Bottleneck in OFGL, specifically targeting the fidelity crisis, and the transferability agnosticism. We introduce High-Fidelity Knowledge Externalization via Fidelity-Driven Proxy Refinement on the client side, and Transferability-Steered Personalized Collaboration on the server side. Extensive experiments validate the effectiveness and superiority of FedHFT in establishing a robust collaborative learning paradigm.
\begin{ack}
This work is supported by National Natural Science Foundation of China under Grant (62361166629, 62501428), the Youth Student Basic Research Project for Doctoral Students of National Natural Science Foundation of China under Grant 623B2080, the China Postdoctoral Science Foundation (2026T190427), Hubei Provincial Natural Science Foundation of China (2025AFB219), the Open Research Fund of Guangdong Laboratory of Artificial Intelligence and Digital Economy (GMLKF-24-10), and the Independent Research Project of the National Key Laboratory of Big Data and Decision (DSJ-ZZKY-2025-02). This work was supported by the National Research Foundation (NRF), Singapore, under Grant ID NRF-PA2025-NTU, and by Nanyang Technological University (NTU). The numerical calculations in this paper have been supported by the super-computing system in the Supercomputing Center of Wuhan University.

\end{ack}

\clearpage
{
\small
\bibliographystyle{plainnat}
\bibliography{references}

@inproceedings{DBLP:conf/nips/0001MO20,
  title={Personalized Federated Learning with Theoretical Guarantees: {A} Model-Agnostic Meta-Learning Approach},
  author={Alireza Fallah and Aryan Mokhtari and Asuman E. Ozdaglar},
  booktitle={NeurIPS},
  year={2020}
}

@inproceedings{DBLP:conf/cvpr/LeeJKOY24,
  title={FedSOL: Stabilized Orthogonal Learning with Proximal Restrictions in Federated Learning},
  author={Gihun Lee and Minchan Jeong and Sangmook Kim and Jaehoon Oh and Se{-}Young Yun},
  booktitle={CVPR},
  pages={12512--12522},
  year={2024}
}

@inproceedings{DBLP:conf/mlsys/LiSZSTS20,
  title={Federated Optimization in Heterogeneous Networks},
  author={Tian Li and Anit Kumar Sahu and Manzil Zaheer and Maziar Sanjabi and Ameet Talwalkar and Virginia Smith},
  booktitle={MLSys},
  year={2020}
}

@inproceedings{DBLP:conf/nips/WangLLJP20,
  title={Tackling the Objective Inconsistency Problem in Heterogeneous Federated Optimization},
  author={Jianyu Wang and Qinghua Liu and Hao Liang and Gauri Joshi and H. Vincent Poor},
  booktitle={NeurIPS},
  year={2020}
}

@inproceedings{DBLP:conf/cvpr/LiHS21,
  title={Model-Contrastive Federated Learning},
  author={Qinbin Li and Bingsheng He and Dawn Song},
  booktitle={CVPR},
  pages={10713--10722},
  year={2021}
}

@inproceedings{DBLP:conf/kdd/ZhangHWSXMG23,
  title={FedCP: Separating Feature Information for Personalized Federated Learning via Conditional Policy},
  author={Jianqing Zhang and Yang Hua and Hao Wang and Tao Song and Zhengui Xue and Ruhui Ma and Haibing Guan},
  booktitle={KDD},
  pages={3249--3261},
  year={2023}
}

@inproceedings{DBLP:conf/iclr/ChenC22,
  title={On Bridging Generic and Personalized Federated Learning for Image Classification},
  author={Hong{-}You Chen and Wei{-}Lun Chao},
  booktitle={ICLR},
  year={2022}
}

@inproceedings{DBLP:conf/ijcai/0010W22,
  title={Adapt to Adaptation: Learning Personalization for Cross-Silo Federated Learning},
  author={Jun Luo and Shandong Wu},
  booktitle={IJCAI},
  pages={2166--2173},
  year={2022}
}

@inproceedings{DBLP:conf/aaai/ZhangHWSXMG23,
  title={FedALA: Adaptive Local Aggregation for Personalized Federated Learning},
  author={Jianqing Zhang and Yang Hua and Hao Wang and Tao Song and Zhengui Xue and Ruhui Ma and Haibing Guan},
  booktitle={AAAI},
  pages={11237--11244},
  year={2023}
}

@inproceedings{DBLP:conf/pkdd/LiZSLS21,
  title={FedPHP: Federated Personalization with Inherited Private Models},
  author={Xin{-}Chun Li and De{-}Chuan Zhan and Yunfeng Shao and Bingshuai Li and Shaoming Song},
  booktitle={ECML PKDD},
  pages={587--602},
  year={2021}
}

@inproceedings{DBLP:conf/iccv/Feng0LXKZ23,
  title={Towards Instance-adaptive Inference for Federated Learning},
  author={Chun{-}Mei Feng and Kai Yu and Nian Liu and Xinxing Xu and Salman Khan and Wangmeng Zuo},
  booktitle={ICCV},
  pages={23230--23239},
  year={2023}
}

@inproceedings{DBLP:conf/iccv/HuTKJ23,
  title={Federated Learning Over Images: Vertical Decompositions and Pre-Trained Backbones Are Difficult to Beat},
  author={Erdong Hu and Yuxin Tang and Anastasios Kyrillidis and Chris Jermaine},
  booktitle={ICCV},
  pages={19328--19339},
  year={2023}
}

@inproceedings{DBLP:conf/iccv/SunM0W023,
  title={FedPerfix: Towards Partial Model Personalization of Vision Transformers in Federated Learning},
  author={Guangyu Sun and Mat{\'{\i}}as Mendieta and Jun Luo and Shandong Wu and Chen Chen},
  booktitle={ICCV},
  pages={4965--4975},
  year={2023}
}

@inproceedings{DBLP:conf/cvpr/WangFKWLG24,
  title={An Aggregation-Free Federated Learning for Tackling Data Heterogeneity},
  author={Yuan Wang and Huazhu Fu and Renuga Kanagavelu and Qingsong Wei and Yong Liu and Rick Siow Mong Goh},
  booktitle={CVPR},
  pages={26223--26232},
  year={2024}
}

@article{DBLP:journals/tpds/ChenLMW22,
  title={FedGraph: Federated Graph Learning With Intelligent Sampling},
  author={Fahao Chen and Peng Li and Toshiaki Miyazaki and Celimuge Wu},
  journal={IEEE TPDS},
  pages={1775--1786},
  year={2022}
}

@inproceedings{DBLP:conf/cvpr/Hayder024,
  title={{DSGG:} Dense Relation Transformer for an End-to-End Scene Graph Generation},
  author={Zeeshan Hayder and Xuming He},
  booktitle={CVPR},
  pages={28317--28326},
  year={2024}
}

@inproceedings{DBLP:conf/cvpr/LiWGY024,
  title={Leveraging Predicate and Triplet Learning for Scene Graph Generation},
  author={Jiankai Li and Yunhong Wang and Xiefan Guo and Ruijie Yang and Weixin Li},
  booktitle={CVPR},
  pages={28369--28379},
  year={2024}
}

@article{DBLP:journals/nn/LiLLQW24,
  title={Contrastive learning of graphs under label noise},
  author={Xianxian Li and Qiyu Li and De Li and Haodong Qian and Jinyan Wang},
  journal={Neural Networks},
  pages={106113},
  year={2024}
}

@article{DBLP:journals/nn/LiWLHL22,
  title={A privacy preservation framework for feedforward-designed convolutional neural networks},
  author={De Li and Jinyan Wang and Qiyu Li and Yuhang Hu and Xianxian Li},
  journal={Neural Networks},
  pages={14--27},
  year={2022}
}

@inproceedings{DBLP:conf/nips/XieMXY21,
  title={Federated Graph Classification over Non-IID Graphs},
  author={Han Xie and Jing Ma and Li Xiong and Carl Yang},
  booktitle={NeurIPS},
  pages={18839--18852},
  year={2021}
}

@inproceedings{DBLP:conf/aaai/TanLL00Z23,
  title={Federated Learning on Non-IID Graphs via Structural Knowledge Sharing},
  author={Yue Tan and Yixin Liu and Guodong Long and Jing Jiang and Qinghua Lu and Chengqi Zhang},
  booktitle={AAAI},
  pages={9953--9961},
  year={2023}
}

@inproceedings{DBLP:conf/nips/TanWHY24,
  title={FedSSP: Federated Graph Learning with Spectral Knowledge and Personalized Preference},
  author={Zihan Tan and Guancheng Wan and Wenke Huang and Mang Ye},
  booktitle={NeurIPS},
  year={2024}
}

@inproceedings{DBLP:conf/aaai/WanHY24,
  title={Federated Graph Learning under Domain Shift with Generalizable Prototypes},
  author={Guancheng Wan and Wenke Huang and Mang Ye},
  booktitle={AAAI},
  pages={15429--15437},
  year={2024}
}

@inproceedings{DBLP:conf/cvpr/TanWHLZ0Y25,
  title={FedSPA: Generalizable Federated Graph Learning under Homophily Heterogeneity},
  author={Zihan Tan and Guancheng Wan and Wenke Huang and He Li and Guibin Zhang and Carl Yang and Mang Ye},
  booktitle={CVPR},
  pages={15464--15475},
  year={2025}
}

@inproceedings{DBLP:conf/cvpr/DuptyDLFGLL24,
  title={Constrained Layout Generation with Factor Graphs},
  author={Mohammed Haroon Dupty and Yanfei Dong and Sicong Leng and Guoji Fu and Yong Liang Goh and Wei Lu and Wee Sun Lee},
  booktitle={CVPR},
  pages={12851--12860},
  year={2024}
}

@inproceedings{DBLP:conf/cvpr/Gan0J00L24,
  title={SignGraph: {A} Sign Sequence is Worth Graphs of Nodes},
  author={Shiwei Gan and Yafeng Yin and Zhiwei Jiang and Hongkai Wen and Lei Xie and Sanglu Lu},
  booktitle={CVPR},
  pages={13470--13479},
  year={2024}
}

@inproceedings{DBLP:conf/cvpr/0005SZZ024,
  title={Tumor Micro-Environment Interactions Guided Graph Learning for Survival Analysis of Human Cancers from Whole-Slide Pathological Images},
  author={Wei Shao and Yangyang Shi and Daoqiang Zhang and Junjie Zhou and Peng Wan},
  booktitle={CVPR},
  pages={11694--11703},
  year={2024}
}

@inproceedings{DBLP:conf/nips/0081C0L0DS022,
  title={{DENSE:} Data-Free One-Shot Federated Learning},
  author={Jie Zhang and Chen Chen and Bo Li and Lingjuan Lyu and Shuang Wu and Shouhong Ding and Chunhua Shen and Chao Wu},
  booktitle={NeurIPS},
  year={2022}
}

@inproceedings{DBLP:conf/iclr/HeinbaughLS23,
  title={Data-Free One-Shot Federated Learning Under Very High Statistical Heterogeneity},
  author={Clare Elizabeth Heinbaugh and Emilio Luz{-}Ricca and Huajie Shao},
  booktitle={ICLR},
  year={2023}
}

@inproceedings{DBLP:conf/aistats/Jhunjhunwala0J24,
  title={FedFisher: Leveraging Fisher Information for One-Shot Federated Learning},
  author={Divyansh Jhunjhunwala and Shiqiang Wang and Gauri Joshi},
  booktitle={AISTATS},
  pages={1612--1620},
  year={2024}
}

@inproceedings{DBLP:conf/aaai/YangSLX24,
  title={Exploring One-Shot Semi-supervised Federated Learning with Pre-trained Diffusion Models},
  author={Mingzhao Yang and Shangchao Su and Bin Li and Xiangyang Xue},
  booktitle={AAAI},
  pages={16325--16333},
  year={2024}
}

@inproceedings{DBLP:conf/icml/QianW0ZW0Y25,
  title={{GHOST:} Generalizable One-Shot Federated Graph Learning with Proxy-Based Topology Knowledge Retention},
  author={Jiaru Qian and Guancheng Wan and Wenke Huang and Guibin Zhang and Yuxin Wu and Bo Du and Mang Ye},
  booktitle={ICML},
  year={2025}
}

@inproceedings{pmlr-v54-mcmahan17a,
  title={{Communication-Efficient Learning of Deep Networks from Decentralized Data}},
  author={McMahan, Brendan and Moore, Eider and Ramage, Daniel and Hampson, Seth and Arcas, Blaise Aguera y},
  booktitle={AISTATS},
  pages={1273--1282},
  year={2017}
}

@inproceedings{DBLP:conf/ijcai/ZhuLWWHL24,
  author       = {Yinlin Zhu and
                  Xunkai Li and
                  Zhengyu Wu and
                  Di Wu and
                  Miao Hu and
                  Rong{-}Hua Li},
  title        = {FedTAD: Topology-aware Data-free Knowledge Distillation for Subgraph
                  Federated Learning},
  booktitle    = {IJCAI},
  pages        = {5716--5724},
  year         = {2024},
 
}

@article{DBLP:journals/pvldb/LiWZZLW23,
  author       = {Xunkai Li and
                  Zhengyu Wu and
                  Wentao Zhang and
                  Yinlin Zhu and
                  Ronghua Li and
                  Guoren Wang},
  title        = {FedGTA: Topology-aware Averaging for Federated Graph Learning},
  journal      = {VLDB},
  pages        = {41--50},
  year         = {2023}
}

@inproceedings{DBLP:conf/ijcai/HuangWY023,
  author       = {Wenke Huang and
                  Guancheng Wan and
                  Mang Ye and
                  Bo Du},
  title        = {Federated Graph Semantic and Structural Learning},
  booktitle    = {IJCAI},
  pages        = {3830--3838},
  year         = {2023}
}

@inproceedings{NEURIPS2024_ba0ad9d1,
  title={One-shot Federated Learning via Synthetic Distiller-Distillate Communication},
  author={Zhang, Junyuan and Liu, Songhua and Wang, Xinchao},
  booktitle={NeurIPS},
  pages={102611--102633},
  year={2024}
}

@inproceedings{wan2026oasis,
  title={{OASIS}: One-Shot Federated Graph Learning via Wasserstein Assisted Knowledge Integration},
  author={Guancheng Wan and Jiaru Qian and Wenke Huang and Qilin Xu and Xianda Guo and Boheng Li and Guibin Zhang and Bo Du and Mang Ye},
  booktitle={NeurIPS},
  year={2026}
}

@inproceedings{DBLP:conf/icml/BaekJJYH23,
  author       = {Jinheon Baek and
                  Wonyong Jeong and
                  Jiongdao Jin and
                  Jaehong Yoon and
                  Sung Ju Hwang},
  title        = {Personalized Subgraph Federated Learning},
  booktitle    = {ICML},
  pages        = {1396--1415},
  year         = {2023}
}

@inproceedings{DBLP:conf/cvpr/HuangY022,
  author       = {Wenke Huang and
                  Mang Ye and
                  Bo Du},
  title        = {Learn from Others and Be Yourself in Heterogeneous Federated Learning},
  booktitle    = {CVPR},
  pages        = {10133--10143},
  year         = {2022}
}

@article{DBLP:journals/pami/HuangYSWLDY24,
  author       = {Wenke Huang and
                  Mang Ye and
                  Zekun Shi and
                  Guancheng Wan and
                  He Li and
                  Bo Du and
                  Qiang Yang},
  title        = {Federated Learning for Generalization, Robustness, Fairness: {A} Survey
                  and Benchmark},
  journal      = {IEEE TPAMI},
  year         = {2024}
}

@ARTICLE{10.3389/fgene.2021.690049,
    
AUTHOR={Zhang, Xiao-Meng  and Liang, Li  and Liu, Lin  and Tang, Ming-Jing },
           
TITLE={Graph Neural Networks and Their Current Applications in Bioinformatics},
          
JOURNAL={Frontiers in Genetics},
  
YEAR={2021}
}

@article{DBLP:journals/isci/0004JWLJJH22,
  author       = {Fei Wu and
                  Xiao{-}Yuan Jing and
                  Pengfei Wei and
                  Chao Lan and
                  Yimu Ji and
                  Guo{-}Ping Jiang and
                  Qinghua Huang},
  title        = {Semi-supervised multi-view graph convolutional networks with application
                  to webpage classification},
  journal      = {Inf. Sci.},
  pages        = {142--154},
  year         = {2022}
}

@article{DBLP:journals/sigkdd/FuZDCL22,
  author       = {Xingbo Fu and
                  Binchi Zhang and
                  Yushun Dong and
                  Chen Chen and
                  Jundong Li},
  title        = {Federated Graph Machine Learning: {A} Survey of Concepts, Techniques,
                  and Applications},
  journal      = {SIGKDD},
  pages        = {32--47},
  year         = {2022}
}

@inproceedings{DBLP:conf/ijcai/LiuT0SJL0J025,
  author       = {Xiang Liu and
                  Zhenheng Tang and
                  Xia Li and
                  Yijun Song and
                  Sijie Ji and
                  Zemin Liu and
                  Bo Han and
                  Linshan Jiang and
                  Jialin Li},
  title        = {One-shot Federated Learning Methods: {A} Practical Guide},
  booktitle    = {IJCAI},
  pages        = {10573--10581},
  year         = {2025}
}

@inproceedings{10.1145/3539597.3570377,
author = {Gao, Yuan and Wang, Xiang and He, Xiangnan and Liu, Zhenguang and Feng, Huamin and Zhang, Yongdong},
title = {Alleviating Structural Distribution Shift in Graph Anomaly Detection},
year = {2023},
booktitle = {WSDM},
pages = {357–365}
}

@article{DBLP:journals/tmc/YeCHCC26,
  author       = {Mang Ye and
                  Yuhang Chen and
                  Wenke Huang and
                  Hui Cai and
                  Laizhong Cui},
  title        = {Towards Generalization Fairness in Federated Learning},
  journal      = {IEEE TMC},
  pages        = {2546--2560},
  year         = {2026}
}

@inproceedings{DBLP:conf/nips/Ma00ZM24,
  author       = {Evelyn Ma and
                  Chao Pan and
                  S. Rasoul Etesami and
                  Han Zhao and
                  Olgica Milenkovic},
  title        = {FedGTST: Boosting Global Transferability of Federated Models via Statistics
                  Tuning},
  booktitle    = {NeurIPS},
  year         = {2024}
}

@inproceedings{DBLP:conf/nips/ZhangSH24,
  author       = {Jianqiao Zhang and
                  Caifeng Shan and
                  Jungong Han},
  title        = {FedGMKD: An Efficient Prototype Federated Learning Framework through
                  Knowledge Distillation and Discrepancy-Aware Aggregation},
  booktitle    = {NeurIPS},
  year         = {2024}
}

@inproceedings{DBLP:conf/ijcai/LiHS21,
  author       = {Qinbin Li and
                  Bingsheng He and
                  Dawn Song},
  title        = {Practical One-Shot Federated Learning for Cross-Silo Setting},
  booktitle    = {IJCAI},
  pages        = {1484--1490},
  year         = {2021}
}

@inproceedings{DBLP:conf/nips/AllouahDGGKP0024,
  author       = {Youssef Allouah and
                  Akash Balasaheb Dhasade and
                  Rachid Guerraoui and
                  Nirupam Gupta and
                  Anne{-}Marie Kermarrec and
                  Rafael Pinot and
                  Rafael Pires and
                  Rishi Sharma},
  title        = {Revisiting Ensembling in One-Shot Federated Learning},
  booktitle    = {NeurIPS},
  year         = {2024}
}

@inproceedings{DBLP:conf/nips/Wang00G024,
  author       = {Haolin Wang and
                  Xuefeng Liu and
                  Jianwei Niu and
                  Wenkai Guo and
                  Shaojie Tang},
  title        = {Why Go Full? Elevating Federated Learning Through Partial Network
                  Updates},
  booktitle    = {NeurIPS},
  year         = {2024}
}

@inproceedings{DBLP:conf/kdd/TanWHYY26,
  author       = {Zihan Tan and
                  Guancheng Wan and
                  Wenke Huang and
                  Bin Yang and
                  Mang Ye},
  title        = {FedPRE: Robust Federated Graph Learning against Topological Corruption},
  booktitle    = {KDD},
  pages        = {1390--1400},
  year         = {2026}
}

@inproceedings{DBLP:conf/icml/ZengHZWW0C25,
  author       = {Hui Zeng and
                  Wenke Huang and
                  Tongqing Zhou and
                  Xinyi Wu and
                  Guancheng Wan and
                  Yingwen Chen and
                  Zhiping Cai},
  title        = {Does One-shot Give the Best Shot? Mitigating Model Inconsistency in
                  One-shot Federated Learning},
  booktitle    = {ICML},
  year         = {2025}
}

@inproceedings{DBLP:conf/nips/YangPW00PW0F24,
  author       = {Ziqi Yang and
                  Zhaopeng Peng and
                  Zihui Wang and
                  Jianzhong Qi and
                  Chaochao Chen and
                  Weike Pan and
                  Chenglu Wen and
                  Cheng Wang and
                  Xiaoliang Fan},
  title        = {Federated Graph Learning for Cross-Domain Recommendation},
  booktitle    = {NeurIPS},
  year         = {2024}
}

@article{DBLP:journals/ir/McCallumNRS00,
  author       = {Andrew Kachites McCallum and
                  Kamal Nigam and
                  Jason Rennie and
                  Kristie Seymore},
  title        = {Automating the Construction of Internet Portals with Machine Learning},
  journal      = {Inf. Retr.},
  pages        = {127--163},
  year         = {2000}
}

@inproceedings{DBLP:conf/dl/GilesBL98,
  author       = {C. Lee Giles and
                  Kurt Dewitt Bollacker and
                  Steve Lawrence},
  title        = {CiteSeer: An Automatic Citation Indexing System},
  booktitle    = {ACM DL},
  pages        = {89--98},
  year         = {1998}
}

@article{canese2013pubmed,
  title={PubMed: the bibliographic database},
  author={Canese, Kathi and Weis, Sarah},
  journal={The NCBI handbook},
  pages={2013},
  year={2013}
}

@inproceedings{DBLP:conf/iclr/PeiWCLY20,
  author       = {Hongbin Pei and
                  Bingzhe Wei and
                  Kevin Chen{-}Chuan Chang and
                  Yu Lei and
                  Bo Yang},
  title        = {Geom-GCN: Geometric Graph Convolutional Networks},
  booktitle    = {ICLR},
  year         = {2020}
}

@article{DBLP:journals/corr/abs-1811-05868,
  author       = {Oleksandr Shchur and
                  Maximilian Mumme and
                  Aleksandar Bojchevski and
                  Stephan G{\"{u}}nnemann},
  title        = {Pitfalls of Graph Neural Network Evaluation},
  journal      = {CoRR},
  year         = {2018}
}

@inproceedings{DBLP:conf/iclr/PlatonovKDBP23,
  author       = {Oleg Platonov and
                  Denis Kuznedelev and
                  Michael Diskin and
                  Artem Babenko and
                  Liudmila Prokhorenkova},
  title        = {A critical look at the evaluation of GNNs under heterophily: Are we
                  really making progress?},
  booktitle    = {ICLR},
  year         = {2023}
}

@inproceedings{DBLP:conf/nips/HuFZDRLCL20,
  author       = {Weihua Hu and
                  Matthias Fey and
                  Marinka Zitnik and
                  Yuxiao Dong and
                  Hongyu Ren and
                  Bowen Liu and
                  Michele Catasta and
                  Jure Leskovec},
  title        = {Open Graph Benchmark: Datasets for Machine Learning on Graphs},
  booktitle    = {NeurIPS},
  year         = {2020}
}

@inproceedings{DBLP:conf/ijcai/JiaoZXTZW25,
  author       = {Pengfei Jiao and
                  Zian Zhou and
                  Meiting Xue and
                  Huijun Tang and
                  Zhidong Zhao and
                  Huaming Wu},
  title        = {FedCCH: Automatic Personalized Graph Federated Learning for Inter-Client
                  and Intra-Client Heterogeneity},
  booktitle    = {IJCAI},
  pages        = {2991--2999},
  year         = {2025}
}

@inproceedings{NEURIPS2025_e2994aef,
 author = {Liao, Dongping and Gao, Xitong and Xu, Cheng-Zhong},
 booktitle = {NeurIPS},
 title = {FLiP: Towards Comprehensive and Reliable Evaluation of Federated Prompt Learning},
 year = {2025}
}

@inproceedings{NEURIPS2025_cdd30bf1,
 author = {Ye, Pukang and Junwei, Luo and Shen, Jiachen and Zhou, Saipan and Dou, Shangmin and Cao, Zhenfu and Yao, Hanzhe and Dong, Xiaolei and Yang, Yunbo},
 booktitle = {NeurIPS},
 pages = {140725--140750},
 title = {FedRW: Efficient Privacy-Preserving Data Reweighting for Enhancing Federated Learning of Language Models},
 year = {2025}
}

@inproceedings{NEURIPS2025_5df817c5,
 author = {Goswami, Dipam and Magistri, Simone and Wang, Kai and Twardowski, Bart\l omiej and Bagdanov, Andrew and van de Weijer, Joost},
 booktitle = {NeurIPS},
 pages = {65081--65115},
 title = {Covariances for Free: Exploiting Mean Distributions for Training-free Federated Learning},
 year = {2025}
}

@inproceedings{NEURIPS2025_3e8f5d29,
 author = {Xu, Kunlun and Feng, Yibo and Li, Jiangmeng and Qi, Yongsheng and Zhou, Jiahuan},
 booktitle = {NeurIPS},
 pages = {44109--44139},
 title = {C\^{}2Prompt: Class-aware Client Knowledge Interaction for Federated Continual Learning},
 year = {2025}
}

@inproceedings{NEURIPS2025_bb203e93,
 author = {Zhang, Haochen and Zheng, Zhong and Xue, Lingzhou},
 booktitle = {NeurIPS},
 pages = {128711--128799},
 title = {Regret-Optimal Q-Learning with Low Cost for Single-Agent and Federated Reinforcement Learning},
 year = {2025}
}

@inproceedings{NEURIPS2025_87ee1bba,
 author = {Yang, Zhiqin and Zhang, Yonggang and Li, Chenxin and Cheung, Yiu-ming and Han, Bo and Yuan, Yixuan},
 booktitle = {NeurIPS},
 pages = {94731--94772},
 title = {FedGPS: Statistical Rectification Against Data Heterogeneity in Federated Learning},
 year = {2025}
}

@inproceedings{NEURIPS2025_6d0c18c6,
 author = {Wei, Ting and Mei, Biao and Lyu, Junliang and Zhang, Renquan and Zhou, Feng and Sun, Yifan},
 booktitle = {NeurIPS},
 pages = {75398--75427},
 title = {Personalized Bayesian Federated Learning with Wasserstein Barycenter Aggregation},
 year = {2025}
}

@inproceedings{NEURIPS2025_74fc5575,
 author = {Ayromlou, Sana and Emerson, David B.},
 booktitle = {NeurIPS},
 pages = {81479--81506},
 title = {Adaptive Latent-Space Constraints in Personalized Federated Learning},
 year = {2025}
}

@inproceedings{NEURIPS2025_f2e61505,
 author = {Motamedi, Azadeh and Kang, Jae-Mo and Kim, Il-Min},
 booktitle = {NeurIPS},
 pages = {166297--166320},
 title = {NormFit: A Lightweight Solution for Few-Shot Federated Learning with Non-IID Data},
 year = {2025}
}

@inproceedings{NEURIPS2025_39540967,
 author = {Ying, Bicheng and Li, Zhe and Yang, Haibo},
 booktitle = {NeurIPS},
 pages = {40156--40201},
 title = {Exact and Linear Convergence for Federated Learning under Arbitrary Client Participation is Attainable},
 year = {2025}
}

@inproceedings{NEURIPS2025_a2582a9f,
 author = {Yang, Danni and Chen, Zhikang and Cui, Sen and Yang, Mengyue and Li, Ding and Wuerkaixi, Abudukelimu and Li, Haoxuan and Ren, Jinke and Gong, Mingming},
 booktitle = {NeurIPS},
 pages = {111897--111926},
 title = {Decentralized Dynamic Cooperation of Personalized Models for Federated Continual Learning},
 year = {2025}
}

@inproceedings{NEURIPS2025_9421261e,
 author = {Wu, Fan and Yan, Xinyu and Liu, Jiabei and Lim, Wei Yang Bryan},
 booktitle = {NeurIPS},
 pages = {102379--102406},
 title = {FedRAM: Federated Reweighting and Aggregation for Multi-Task Learning},
 year = {2025}
}

@inproceedings{NEURIPS2025_16e71d1a,
 author = {Wu, Jiahao and Hu, Ming and Yang, Yanxin and Xie, Xiaofei and Chen, ZeKai and Song, Chenyu and Chen, Mingsong},
 booktitle = {NeurIPS},
 pages = {15502--15528},
 title = {Rising from Ashes: Generalized Federated Learning via Dynamic Parameter Reset},
 year = {2025}
}

@inproceedings{NEURIPS2025_659afd14,
 author = {Wan, Frank and Ran, Fengyuan and Zhang, Ruikang and Huang, Wenke and Rong, Xuankun and Zhang, Guibin and Wu, Yuxin and Du, Bo and Ye, Mang},
 booktitle = {NeurIPS},
 pages = {70280--70304},
 title = {MOTION: Multi-Sculpt Evolutionary Coarsening for Federated Continual Graph Learning},
 year = {2025}
}

@inproceedings{NEURIPS2025_5ee643ae,
 author = {Wan, Frank and Shang, Xiaoran and Wu, Yuxin and Zhang, Guibin and Bi, Jinhe and Zheng, Liangtao and Lin, Xin and Liu, Yue and Ma, Yanbiao and Huang, Wenke and Du, Bo},
 booktitle = {NeurIPS},
 pages = {65655--65680},
 title = {HYPERION: Fine-Grained Hypersphere Alignment for Robust Federated Graph Learning},
 year = {2025}
}

@inproceedings{NEURIPS2025_29c71a71,
 author = {Zuo, Shiyuan and Yan, Xingrun and Fan, Rongfei and Shen, Li and Zhao, Puning and Xu, Jie and Hu, Han},
 booktitle = {NeurIPS},
 pages = {28998--29035},
 title = {Efficient Federated Learning against Byzantine Attacks and Data Heterogeneity via Aggregating Normalized Gradients},
 year = {2025}
}

@inproceedings{NEURIPS2025_66808849,
 author = {Wan, Frank and Cheng, Xu and Liu, Run and Huang, Wenke and Shi, Zitong and Jin, Pinyi and Zhang, Guibin and Du, Bo and Ye, Mang},
 booktitle = {NeurIPS},
 pages = {70587--70611},
 title = {Multi-order Orchestrated Curriculum Distillation for Model-Heterogeneous Federated Graph Learning},
 year = {2025}
}

@inproceedings{NEURIPS2025_7c818dd4,
 author = {Jia, Yuqi and Fang, Minghong and Gong, Neil},
 booktitle = {NeurIPS},
 pages = {86116--86145},
 title = {Competitive Advantage Attacks to Decentralized Federated Learning},
 year = {2025}
}

@inproceedings{NEURIPS2025_a924b717,
 author = {Jia, Yuqi and Fang, Minghong and Liu, Hongbin and Zhang, Jinghuai and Gong, Neil},
 booktitle = {NeurIPS},
 pages = {116470--116497},
 title = {Tracing Back the Malicious Clients in Poisoning Attacks to Federated Learning},
 year = {2025}
}

@inproceedings{NEURIPS2025_dcc0ac74,
 author = {Wu, Siyuan and Jia, Yongzhe and Xiang, Haolong and Xu, Xiaolong and Zhang, Xuyun and Qi, Lianyong and Dou, Wanchun},
 booktitle = {NeurIPS},
 pages = {150102--150131},
 title = {A Fair Federated Learning Method for Handling Client Participation Probability Inconsistencies in Heterogeneous Environments},
 year = {2025}
}

@inproceedings{NEURIPS2025_3327377e,
 author = {Zhang, Letian and Chen, Bo and Bian, Jieming and Wang, Lei and Xu, Jie},
 booktitle = {NeurIPS},
 pages = {35665--35696},
 title = {FedEL: Federated Elastic Learning for Heterogeneous Devices},
 year = {2025}
}

@inproceedings{NEURIPS2025_a0f30168,
 author = {Pelikan, Martin and Azam, Shams and Feldman, Vitaly and Silovsky, Jan and Talwar, Kunal and Brinton, Christopher and Likhomanenko, Tatiana},
 booktitle = {NeurIPS},
 pages = {110958--111020},
 title = {Enabling Differentially Private Federated Learning for Speech Recognition: Benchmarks, Adaptive Optimizers, and Gradient Clipping},
 year = {2025}
}

@inproceedings{NEURIPS2025_aaec5dee,
 author = {Wang, Lingren and Tu, Wenxuan and Wang, Jiaxin and Wang, Xiong and Cheng, Jieren and Liu, Jingxin},
 booktitle = {NeurIPS},
 pages = {117917--117940},
 title = {FedIGL: Federated Invariant Graph Learning for Non-IID Graphs},
 year = {2025}
}

@inproceedings{NEURIPS2025_d8ace30c,
 author = {Man, Tianxing and Bai, Yu and Wang, Ganyu and Fang, Jinjie and Fang, Haoran and Gu, Bin and Chang, Yi},
 booktitle = {NeurIPS},
 pages = {147561--147593},
 title = {Accelerated Vertical Federated Adversarial Learning through Decoupling Layer-Wise Dependencies},
 year = {2025}
}

@inproceedings{NEURIPS2025_57fbe68c,
 author = {Yan, Gang and Yang, Sikai and Du, Wan},
 booktitle = {NeurIPS},
 pages = {61155--61178},
 title = {FedRACE: A Hierarchical and Statistical Framework for Robust Federated Learning},
 year = {2025}
}

@inproceedings{NEURIPS2025_e6a11b61,
 author = {BONNERJEE, SOHAM and Karmakar, Sayar and Wu, Wei Biao},
 booktitle = {NeurIPS},
 pages = {157224--157270},
 title = {Sharp Gaussian approximations for Decentralized Federated Learning},
 year = {2025}
}

@inproceedings{NEURIPS2025_7d3f1540,
 author = {Li, Yunbo and Gui, Jiaping and Deng, Zhihang and Meng, Fanchao and Wu, Yue},
 booktitle = {NeurIPS},
 pages = {86502--86550},
 title = {FedQS: Optimizing Gradient and Model Aggregation for Semi-Asynchronous Federated Learning},
 year = {2025}
}

@inproceedings{NEURIPS2025_7ccae68f,
 author = {Kim, Jisoo and Kang, Sungmin and Lee, Sunwoo},
 booktitle = {NeurIPS},
 pages = {86242--86273},
 title = {Layer-wise Update Aggregation with Recycling for Communication-Efficient Federated Learning},
 year = {2025}
}

@inproceedings{NEURIPS2025_d611d06e,
 author = {Qi, Zhuang and Tang, Ying-Peng and Meng, Lei and Yu, Han and Li, Xiaoxiao and Meng, Xiangxu},
 booktitle = {NeurIPS},
 publisher = {Curran Associates, Inc.},
 title = {Class-wise Balancing Data Replay for Federated Class-Incremental Learning},
 year = {2025}
}

@inproceedings{NEURIPS2025_76ee49ee,
 author = {Tang, Xiaoli and Yu, Han and Li, Xiaoxiao},
 booktitle = {NeurIPS},
 pages = {82666--82684},
 title = {A Reinforcement Learning-based Bidding Strategy for Data Consumers in Auction-based Federated Learning},
 year = {2025}
}

@inproceedings{NEURIPS2025_ba0dd552,
 author = {HUYNH, Khiem and Egan, Malcolm and Neglia, Giovanni and Gorce, Jean-Marie},
 booktitle = {NeurIPS},
 pages = {128137--128189},
 title = {Streaming Federated Learning with Markovian Data},
 year = {2025}
}

@inproceedings{NEURIPS2025_67496dfa,
 author = {Fenoglio, Dario and Li, Mohan and Barbiero, Pietro and Lane, Nicholas and Langheinrich, Marc and Gjoreski, Martin},
 booktitle = {NeurIPS},
 pages = {71229--71292},
 title = {FLUX: Efficient Descriptor-Driven Clustered Federated Learning under Arbitrary Distribution Shifts},
 year = {2025}
}

@inproceedings{NEURIPS2025_f6913c4f,
 author = {Li, Xiang and Wang, Chendi and Su, Buxin and Long, Qi and Su, Weijie},
 booktitle = {NeurIPS},
 pages = {168588--168624},
 title = {Mitigating the Privacy\textendash Utility Trade-off in Decentralized Federated Learning via f-Differential Privacy},
 year = {2025}
}

@inproceedings{NEURIPS2025_2d945a80,
 author = {Liang, Dandan and Zhang, Jianing and Chen, Evan and Li, Zhe and Li, Rui and Yang, Haibo},
 booktitle = {NeurIPS},
 pages = {31634--31673},
 title = {Towards Straggler-Resilient Split Federated Learning: An Unbalanced Update Approach},
 year = {2025}
}

@inproceedings{NEURIPS2025_6a9b9b04,
 author = {Wu, Xinghao and Liu, Xuefeng and Niu, Jianwei and Zhu, Guogang and Shi, Mingjia and Tang, Shaojie and Yuan, Jing},
 booktitle = {NeurIPS},
 pages = {73655--73691},
 title = {Tackling Feature-Classifier Mismatch in Federated Learning via Prompt-Driven Feature Transformation},
 year = {2025}
}

@inproceedings{NEURIPS2025_721dbcfe,
 author = {Li, Yichen and Wang, Yuying and Dong, Jiahua and Wang, Haozhao and Qi, Yining and Zhang, Rui and Li, Ruixuan},
 booktitle = {NeurIPS},
 pages = {79493--79518},
 title = {Resource-Constrained Federated Continual Learning: What Does Matter?},
 year = {2025}
}

@inproceedings{NEURIPS2025_8c4c80d3,
 author = {Fu, Lele and Liao, Tianchi and Huang, Sheng and Deng, Bowen and zhangchuanfu and Pan, Shirui and Chen, Chuan},
 booktitle = {NeurIPS},
 pages = {97431--97457},
 title = {Unsupervised Federated Graph Learning},
 year = {2025}
}

@inproceedings{NEURIPS2025_0e614c59,
 author = {Huang, Sheng and Fu, Lele and Ye, Fanghua and Liao, Tianchi and Deng, Bowen and zhangchuanfu and Chen, Chuan},
 booktitle = {NeurIPS},
 pages = {9968--10002},
 title = {Soft-consensual Federated Learning for Data Heterogeneity via Multiple Paths},
 year = {2025}
}

@inproceedings{NEURIPS2025_06d2e62a,
 author = {Li, Tianle and Huang, Yongzhi and Jiang, Linshan and Liu, Chang and Xie, Qipeng and Du, Wenfeng and Wang, Lu and Wu, Kaishun},
 booktitle = {NeurIPS},
 pages = {4425--4462},
 title = {FedWMSAM: Fast and Flat Federated Learning via Weighted Momentum and Sharpness-Aware Minimization},
 year = {2025}
}

@inproceedings{NEURIPS2025_cfcb1905,
 author = {Fang, Xiuwen and Ye, Mang},
 booktitle = {NeurIPS},
 pages = {141777--141800},
 title = {Towards Robust Parameter-Efficient Fine-Tuning for Federated Learning},
 year = {2025}
}

@inproceedings{NEURIPS2025_4a4d5fca,
 author = {Bo, Weihao and Sun, Yanpeng and Wang, Yu and Zhang, Xinyu and Li, Zechao},
 booktitle = {NeurIPS},
 pages = {51792--51826},
 title = {FedMGP: Personalized Federated Learning with Multi-Group Text-Visual Prompts},
 year = {2025}
}

@inproceedings{NEURIPS2025_adb02256,
 author = {Xue, Jiaqi and Kumar, Mayank and Shang, Yuzhang and Gao, Shangqian and Ning, Rui and Zheng, Mengxin and Jiang, Xiaoqian and Lou, Qian},
 booktitle = {NeurIPS},
 pages = {119806--119829},
 title = {DictPFL: Efficient and Private Federated Learning on Encrypted Gradients},
 year = {2025}
}

@inproceedings{NEURIPS2025_29978617,
 author = {Nguyen, Duong and Hoang, Nghia and Huynh, Thanh Trung and Nguyen, Quoc Viet Hung and Nguyen, Phi Le},
 booktitle = {NeurIPS},
 pages = {28808--28845},
 title = {Learning Reconfigurable Representations for Multimodal Federated Learning with Missing Data},
 year = {2025}
}

@inproceedings{wan2025energy,
  title={Energy-based backdoor defense against federated graph learning},
  author={Wan, Guancheng and Shi, Zitong and Huang, Wenke and Zhang, Guibin and Tao, Dacheng and Ye, Mang},
  booktitle={ICLR},
  year={2025}
}

@article{shi2025privacy,
  title={Privacy-enhancing paradigms within federated multi-agent systems},
  author={Shi, Zitong and Wan, Guancheng and Huang, Wenke and Zhang, Guibin and Shao, Jiawei and Ye, Mang and Yang, Carl},
  journal={arXiv preprint arXiv:2503.08175},
  year={2025}
}

@inproceedings{shi2025eagles,
  title={EAGLES: Towards Effective, Efficient, and Economical Federated Graph Learning via Unified Sparsification.},
  author={Shi, Zitong and Wan, Guancheng and Huang, Wenke and Zhang, Guibin and Li, He and Yang, Carl and Ye, Mang},
  booktitle={ICML},
  year={2025}
}

@inproceedings{shi2026towards,
  title={Towards robust text-attributed federated graph learning: multimodal threats and defense},
  author={Shi, Zitong and Wan, Guancheng and Huang, Wenke and Wu, Yuxin and Zhang, Quan and Ye, Mang},
  booktitle={AAAI},
  year={2026}
}

@article{shi2026don,
  title={Don’t Forget the Enjoin: FocalLoRA for Instruction Hierarchical Alignment in Large Language Models},
  author={Shi, Zitong and Wan, Frank and Wang, Haixin and Li, Ruoyan and Huang, Zijie and Zhao, Wanjia and Xiao, Yijia and Luo, Xiao and Yang, Carl and Sun, Yizhou and others},
  journal={NeurIPS},
  year={2026}
}

@article{tan2025s2fgl,
title={S2fgl: Spatial spectral federated graph learning},
author={Tan, Zihan and Huang, Suyuan and Wan, Guancheng and Huang, Wenke and Li, He and Ye, Mang},
journal={arXiv preprint arXiv:2507.02409},
year={2025}
}

@inproceedings{guo2026fedfst,
title={FedFST: Mitigating Spectral Catastrophic Forgetting in Federated Graph Continual Learning},
author={Guo, Hanyao and Tan, Zihan and Huang, Wenke and Yang, Bin and Ye, Mang},
booktitle={KDD},
year={2026}
}

@inproceedings{liu2026fedsdr,
title={FedSDR: Federated Graph Learning with Structural Noise Detection and Reconstruction},
author={Liu, Jiaqi and Tan, Zihan and Wan, Guancheng and Huang, Wenke and Li, He and Ye, Mang},
booktitle={CVPR},
year={2026}
}

@inproceedings{jia2026consistency,
  title     = {Consistency-Verified Backdoor Defense for Federated Graph Learning via Cross-Layer Drift},
  author    = {Jia, Yanwen and Zhang, Yinuo and Shi, Zitong and Wu, Yuxin and Tan, Zihan and Rong, Xuankun and Zheng, Liangtao and Huang, Wenke and Ye, Mang},
  booktitle = {NeurIPS},
  year      = {2026}
}

@inproceedings{wang2026e2gen,
  title     = {{E}$^2${Gen}: Evidential Energy-Based Generation for Fair Federated Graph Learning},
  author    = {Wang, Jingbo and Shi, Zitong and Wu, Yuxin and Tan, Zihan and Lin, Qiqi and Rong, Xuankun and Zheng, Liangtao and Huang, Wenke and Ye, Mang},
  booktitle = {NeurIPS},
  year      = {2026}
}

@article{wu2026swepolyvisionbenchmarkingcrossimageabductive,
title={SWE-PolyVision: Benchmarking Cross-Image Abductive Reasoning for Repository-Level Software Engineering},
author={Wu, Jiajun and Sun, Leixin and Tan, Zihan and Liu, Yitao and Li, Shuo and Qian, Jiaru and Wu, Yuxin and Quan, Shanghaoran and Zhao, Chuangxin and Liao, Yangxu and Liu, Yang and Chong, Bin and Wan, Guancheng},
journal={arXiv preprint arXiv:2609.29754},
year={2026}
}

@article{wu2026sweprometheusmeasuringengineeringgovernance,
title={SWE-Prometheus: Measuring Engineering Governance Improvements in Real-World Repositories},
author={Wu, Jiajun and Sun, Leixin and Tan, Zihan and Liu, Yitao and Li, Shuo and Qian, Jiaru and Wu, Yuxin and Quan, Shanghaoran and Zhao, Chuangxin and Liao, Yangxu and Liu, Yang and Chong, Bin and Wan, Guancheng},
journal={arXiv preprint arXiv:2609.29465},
year={2026}
}

@article{tan2026metarsirsi2metarecursive,
title={MetaRSI / RSI2: A Meta-Recursive Self-Improving System for Recursive Self-Improving Systems Themselves},
author={Tan, Zihan and Sun, Leixin and Shi, Zitong and Liu, Yitao and Wu, Jiajun and Brooks, Nathaniel and Qian, Jiaru and Shang, Xiaoran and Huang, Suyuan and Ding, Yi and Liao, Yangxu and Li, Mukai and Sun, Qiushi and Liu, Shudong and Rong, Xuankun and Yu, Xiaohang and Chen, Zhuo and Geng, Hejia and Li, Chenxin and Wang, Aozhou and Tu, Zengji and Tang, Robert and Zhan, Yuxin and Jiang, Eric and Wu, Yuxin and Zhang, Jianqing and Liang, Xiao and Wu, Fang and Zhang, Haochi and Marlow, Alexander and Wan, Guancheng},
journal={arXiv preprint arXiv:2609.06396},
year={2026}
}
}
\clearpage
\appendix
\section{Notations}
We present a comprehensive review of the commonly used notations and their definitions in Tab.\ref{tab:notations}.

\begin{table*}[!h]
\centering
\caption{Notation and Definitions}
\label{tab:notations}
\begin{tabular}{l l}
\toprule
\textbf{Notation} & \textbf{Definition} \\ 
\midrule
$K$ & The number of clients. \\
$k, i, j$ & The client indices. \\
$N^k$ & The number of nodes on client $k$. \\
$C$ & The total number of classes. \\
$d^k$ & The feature dimension of client $k$. \\
$f_{\theta}$ & The initial global model. \\
$\Theta$ & The set of model parameters. \\
$\mathbf{x}_b^k, y_b^k$ & The real node features and labels. \\
$\tilde{\mathbf{x}}_b^k, \tilde{y}_b^k$ & The proxy node features and labels. \\
$\mathcal{Y}^k$ & The set of valid classes on client $k$. \\ 
$B, B_p$ & The real and proxy batch sizes. \\
$\mathbf{F}_{r}^{k}(\theta)$ & The Fisher statistic on real data. \\
$\mathbf{F}_{p}^{k}(\theta)$ & The Fisher statistic on proxy data. \\
$\mathbf{T}^k$ & The local transition matrix of client $k$. \\
$\mathbf{v}^{k,(t)}$ & The node importance vector at iteration $t$. \\
$v_i^k$ & The topology importance score of node $i$. \\
$\alpha$ & The PageRank damping factor. \\
$\mathbf{p}_c^k$ & The topology-aware class prior. \\
$\mathcal{V}_c^k$ & The node set of class $c$ on client $k$. \\
$\mathcal{L}_{\text{probe}}$ & The probe loss. \\
$\mathbf{g}^k$ & The channel salience vector. \\
$M^k$ & The number of selected informative channels. \\
$\mathcal{I}^k$ & The index set of top-$M$ channels. \\
$\widetilde{\mathbf{X}}^k$ & The generated node feature matrix. \\
$\mathbf{P}_{+}^k, \mathbf{P}_{-}^k$ & The positive and negative prototype matrices. \\
$q^k, p_+^k, p_-^k$ & The softmax distributions on selected channels. \\
$\lambda_{\text{sem}}$ & The trade-off hyperparameter. \\
$\mathcal{L}_{\text{Fis}}^{k}$ & The Fisher matching objective. \\
$\mathcal{L}_{\text{sem}}^{k}$ & The semantic alignment objective. \\ 
$\mathbf{P}^k$ & The uploaded prototype matrix of client $k$. \\
$\mathbf{l}^k[c]$ & The NLL of prototype $\mathbf{p}_c^k$. \\
$\mathbf{L}$ & The stacked loss matrix. \\
$\mathbf{l}^{i,(\cap)}$ & The loss vector on overlapping classes. \\
$T_{ij}$ & The estimated transferability score. \\
$\mathbf{T}$ & The transferability matrix. \\
$W_{ij}$ & The normalized collaboration weight. \\
$\mathbf{W}$ & The personalized collaboration weight matrix. \\
$\tau$ & The temperature hyperparameter. \\
$\mathcal{L}_{\text{sup}}^{(i \leftarrow j)}$ & The supervised loss evaluated on the proxy graph. \\
$\mathcal{L}_i$ & The personalized integration loss. \\ 
\bottomrule
\end{tabular}
\end{table*}

\section{Fisher matching} \label{Fisher_matching}

Parameter sensitivity in our framework is quantified by a Fisher-guided statistic. Let \(\mathcal{G}\) denote the local graph data distribution and \(\theta\) denote the model parameters. The classical Fisher Information Matrix (FIM) is defined as:
\begin{equation}
\mathbf{I}(\theta) = \mathbb{E}_{(x,y)\sim \mathcal{G}} \!\left[ \nabla_{\theta}\log p_{\theta}(y\mid x)\, \nabla_{\theta}\log p_{\theta}(y\mid x)^{\top} \right].
\end{equation}

The FIM inherently captures the curvature of the log-likelihood function in the parameter space. Crucially, it provides a fundamental connection between parameter perturbations and model output variations. Specifically, the expected Kullback-Leibler (KL) divergence between the original distribution and a distribution perturbed by a small parameter shift \(\delta\) can be bounded via the second-order Taylor expansion:
\begin{equation}
\mathbb{E}_{\mathcal{G}} \left[ D_{\mathrm{KL}}(p(\mathcal{G} \mid \theta) \parallel p(\mathcal{G} \mid \theta+\delta)) \right] \approx \frac{1}{2}\delta^{\top}\mathbf{I}(\theta)\delta.
\end{equation}
This property is highly desirable in our one-shot setting, as it theoretically bounds the output divergence caused by optimization trajectory deviations. 

However, directly computing \(\mathbf{I}(\theta) \in \mathbb{R}^{|\theta|\times|\theta|}\), or even its standard diagonal approximation \(\mathbb{E}[(\nabla_{\theta}\log p)^2]\), is computationally prohibitive for GNNs. The message-passing mechanism structurally entangles node representations, making the acquisition of independent sample-level gradients computationally intractable without incurring catastrophic memory overheads. 

To maintain tractability under the strict resource constraints of one-shot federated learning, we utilize a mini-batch empirical Fisher surrogate. Concretely, for each parameter coordinate \(\theta_t\), we compute the batch-averaged score gradients on real data and proxy data, and square them element-wise to form our sensitivity profiles \(F_{r,t}^{k}\) and \(F_{p,t}^{k}\). The gradient matching objective is formulated as:
\begin{equation}
\mathcal{L}_{\mathrm{Fis}}^{k} = \sum_{t\in\Theta} \left( F_{p,t}^{k}-F_{r,t}^{k} \right)^2.
\end{equation}

Building upon the traditional Fisher formulation, \(F_{r}^{k}\) and \(F_{p}^{k}\) serve as optimized, memory-efficient sensitivity surrogates tailored for graph data. Rather than blindly aligning features, matching \(F_{p}^{k}\) to \(F_{r}^{k}\) explicitly forces the proxy data to induce similar parameter-update tendencies as the real data. This effectively minimizes the trajectory-level gradient-response mismatch and restricts the posterior parameter divergence before the single-round upload.

\section{Sensitivity Analysis of Top-$M$.}\label{topm}

\begin{wrapfigure}{r}{0.45\textwidth}
    \centering
    \includegraphics[width=\linewidth, trim={1.5cm 1.5cm 1.5cm 1.5cm}, clip]{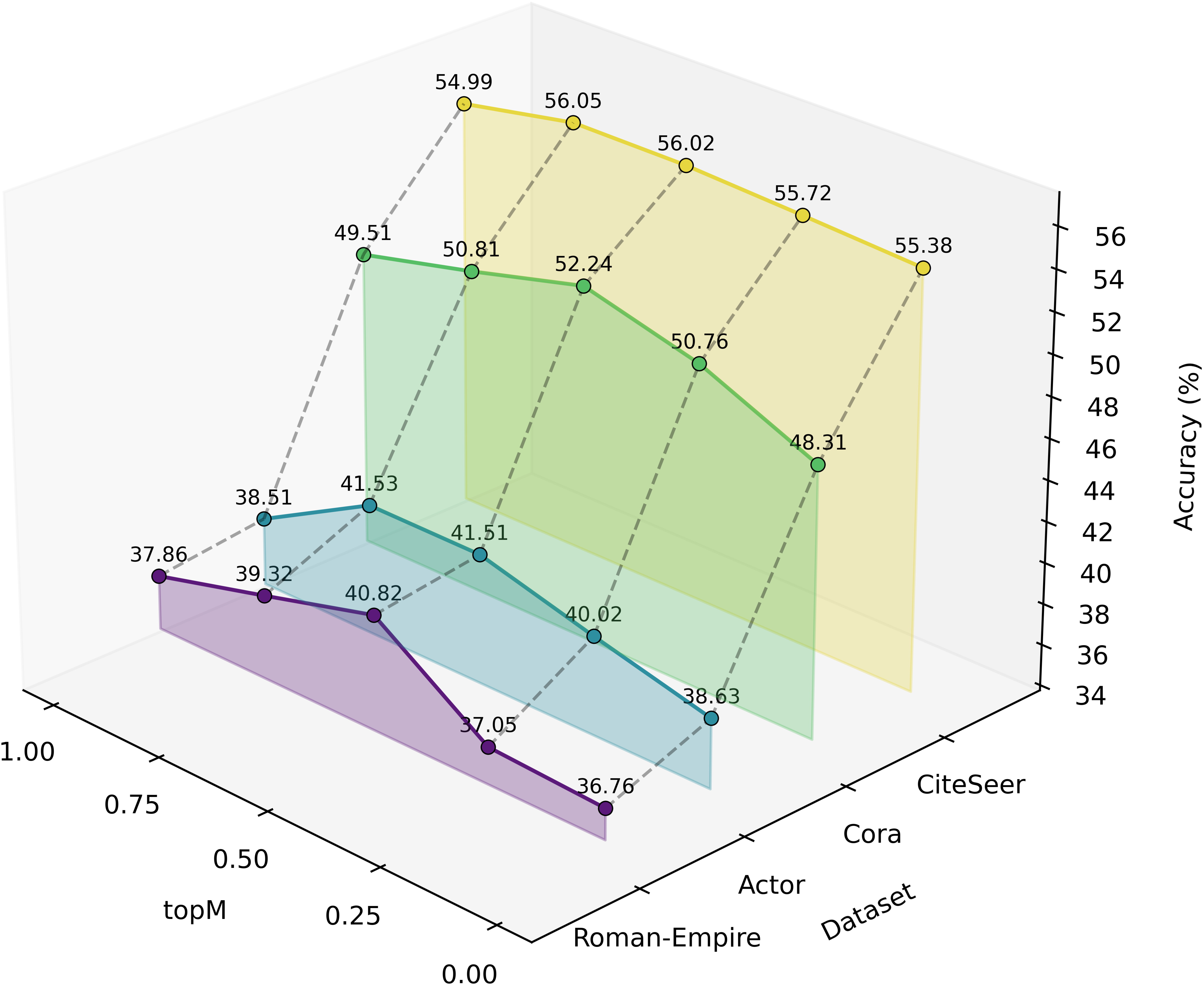}
    \caption{Sensitivity analysis of the Top-$M$ selection ratio.}
    \label{fig:topM_sensitivity}
\end{wrapfigure}
To evaluate the sensitivity of FedHFT to the Top-$M$ selection ratio, we conduct a systematic comparison within the range of $M \in \{0, 0.25, 0.5, 0.75, 1\}$ and report the average test accuracy across multiple datasets, including Actor, CiteSeer, Cora and Roman-Empire. As illustrated in Figure~\ref{fig:topM_sensitivity}, the model generally exhibits a trend where a moderate $M$ yields superior results, whereas excessively small or large values lead to performance degradation. Specifically, as $M$ increases from $0$ to the range of $0.5 \sim 0.75$, the performance on most datasets steadily improves; however, when $M$ further increases to $1$, the accuracy experiences a general decline. This phenomenon indicates that forcing all dimensions to strictly align with class prototypes tends to entangle useful categorical cues with noisy channels, thereby impairing model performance. Consequently, either omitting the filtering process entirely ($M=1$) or applying excessively strict filtering ($M=0$) weakens the model's capabilities. Considering the optimal performance and stability across different datasets, the model demonstrates the most robust behavior around $M=0.5$. Therefore, we adopt $M=0.5$ as the default configuration for FedHFT.

\section{transferability--accuracy correlation}\label{transfer}
\begin{wrapfigure}{r}{0.45\textwidth} 
    \centering
    \includegraphics[width=0.48\linewidth]{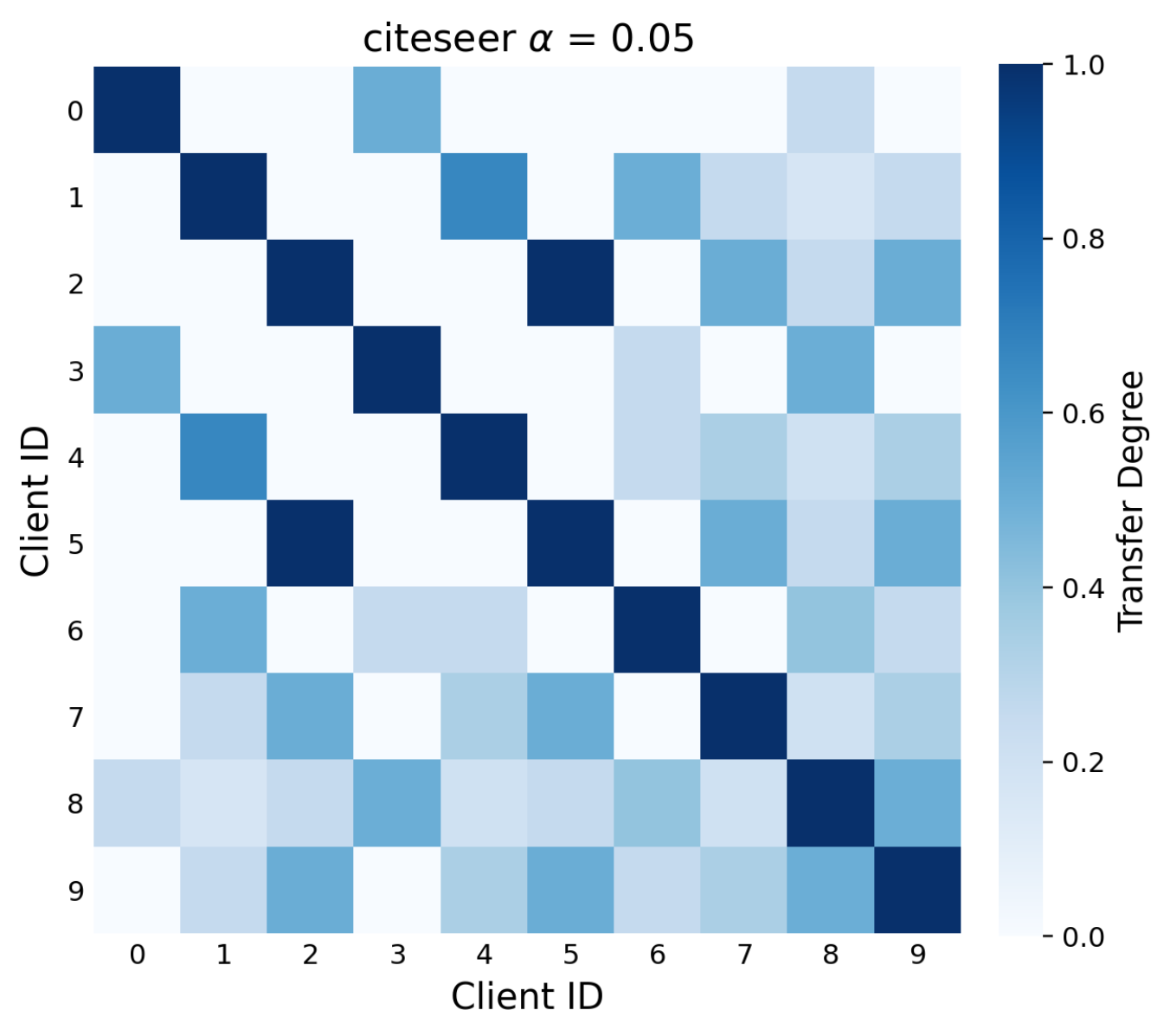}
    \hfill 
    \includegraphics[width=0.48\linewidth]{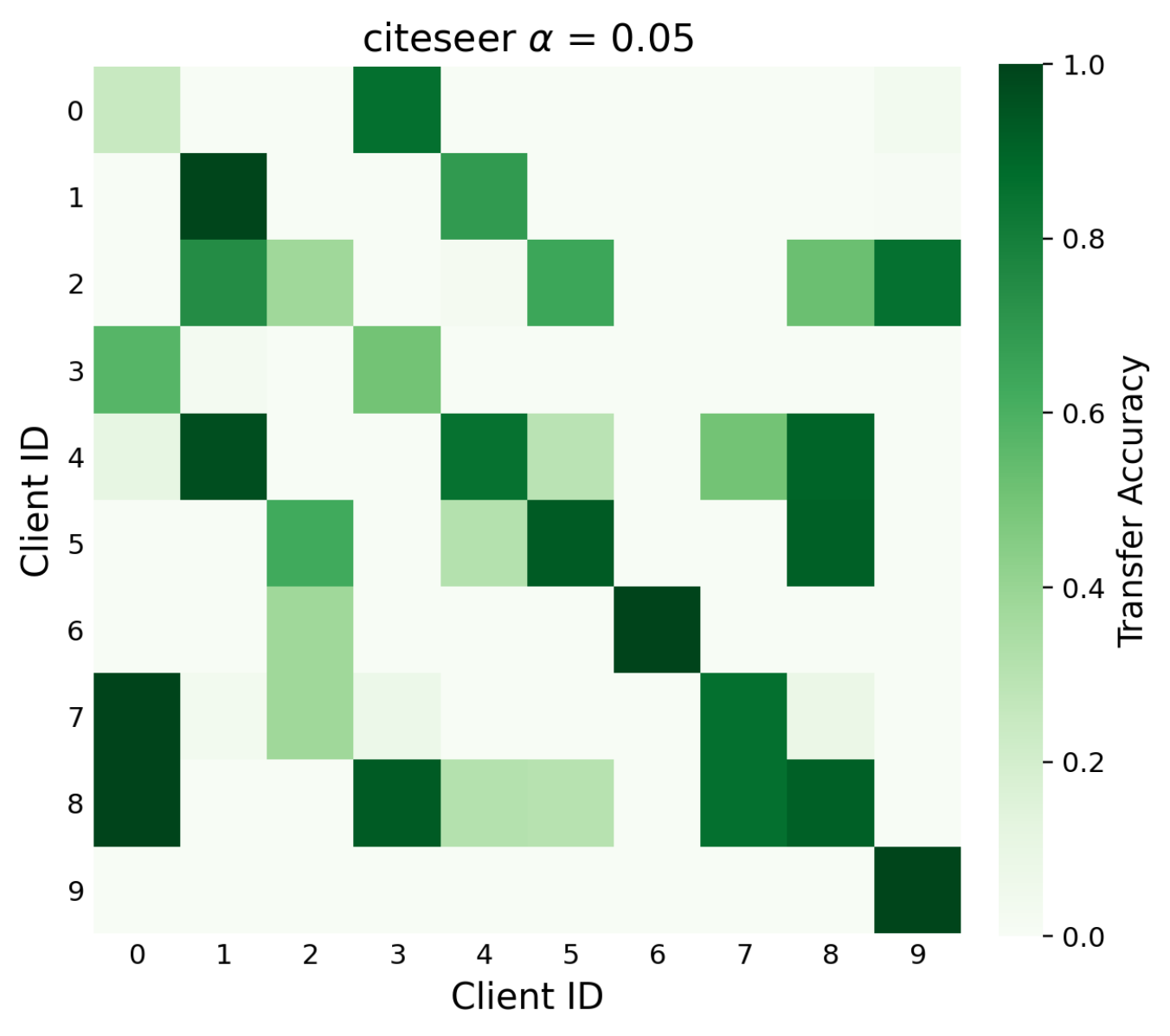}
    \caption{Correlation between estimated transferability and realized transfer accuracy.} 
    \label{fig:transfer_correlation}
\end{wrapfigure}
Our federated framework explicitly routes inter-client knowledge through prototype-based transfer weights and realizes the assigned migration in personalized graph networks. On label-skewed Citeseer at $\alpha=0.05$, we quantify transfer strength with the pairwise transfer degree matrix and measure realized migration with the cross-client transfer accuracy matrix $A_{ij}$, where client $i$'s personalized model is evaluated on client $j$'s synthesized proxy under $j$'s test mask. The result in Fig.\ref{fig:transfer_correlation} confirms that our method not only estimates migration reliably but also converts it into strong personalized performance.

\section{Dataset Details} \label{Dataset_details}
We evaluate our method on seven real-world graph datasets: \textbf{Cora, CiteSeer, PubMed, Actor, Amazon-Photo, Roman-Empire, and OGBN-Arxiv}. According to their homophily characteristics, we group them into \textbf{homophilous} and \textbf{heterophilous} datasets. The key statistics
of these datasets are summarized in table~\ref{tab:dataset_statistics}. The details are as follows:

\begin{itemize}
    \item \textbf{Homophilous datasets (Cora, CiteSeer, PubMed, Amazon-Photo, OGBN-Arxiv).}
    These datasets exhibit strong label consistency among connected nodes and are widely used to evaluate GNNs under homophily-favorable conditions. Cora is a citation network where nodes are papers, edges are citation links, and node features are derived from paper text for topic classification. CiteSeer is also a citation graph of scientific publications, with sparse high-dimensional text features and multiple research topic labels. PubMed is a large biomedical citation network in which papers are represented by TF-IDF-like text features and classified into disease-related categories. Amazon-Photo is a co-purchase graph where nodes are products in the photo category, edges connect frequently co-purchased items, and labels indicate product categories. OGBN-Arxiv is a large-scale directed citation network from arXiv, where node features are derived from paper titles and abstracts, and labels correspond to subject areas.

    \item \textbf{Heterophilous datasets (Actor, Roman-Empire).}
    These datasets contain many edges connecting nodes with different labels, making them suitable for testing model robustness under heterophily. Actor is a co-occurrence network of actors from Wikipedia pages, where node features are bag-of-words attributes and neighboring nodes often belong to different classes. Roman-Empire is a heterophilous benchmark graph with low local label agreement, designed to challenge message passing methods that rely on homophily.
\end{itemize}

\begin{table}[htbp]
\centering
\caption{Statistics of datasets used in experiments.}
\renewcommand{\arraystretch}{1.1}
\begin{tabular}{c||cccc}
\hline
\rowcolor[gray]{0.9}
\textbf{Dataset} & \textbf{\#Nodes} & \textbf{\#Edges} & \textbf{\#Classes} & \textbf{\#Features} \\
\hline
Cora & 2,708 & 5,278 & 7 & 1,433 \\
CiteSeer & 3,327 & 4,552 & 6 & 3,703 \\
PubMed & 19,717 & 44,324 & 3 & 500 \\
Amz-Photo & 7,650 & 287,326 & 8 & 745 \\
OGBN-Arxiv & 169,343 & 1,166,243 & 40 & 128 \\
Actor & 7,600 & 30,019 & 5 & 932 \\
Roman-Empire & 22,622 & 65,854 & 18 & 300 \\
\hline
\end{tabular}
\label{tab:dataset_statistics}
\end{table}

\section{Baseline Details}\label{baseline}
This section provides a comprehensive overview of the baseline approaches employed in our study. To systematically evaluate the proposed FedHFT, we compare it against 14 representative baselines spanning four distinct categories: traditional FL, FGL, OFL, and OFGL. Under our experimental settings, traditional FL and FGL methods are included to demonstrate the performance collapse of multi-round models when suffering from the Single-Transmission Bottleneck.
\begin{itemize}
    \item \textbf{FedAvg} [AISTATS17]. FedAvg introduces a foundational decentralized learning paradigm that trains a shared global model while keeping data localized on client devices. It iteratively aggregates locally computed client updates instead of centralized training. This mechanism addresses privacy constraints and remains robust to unbalanced non-IID data. By reducing continuous server-client synchronization needs, FedAvg establishes a practical and communication-efficient federated learning framework.
    
    \item \textbf{FedProx} [MLSys20]. FedProx addresses systems and statistical heterogeneity in realistic federated networks. As a generalization of FedAvg, it allows clients to perform a variable amount of local work before aggregation under device-level constraints. It also restricts local updates to prevent excessive drift from the global model. This refinement yields more stable, robust, and accurate convergence in highly heterogeneous environments.

    \item \textbf{FedNova} [NeurIPS20]. FedNova improves federated optimization by correcting objective inconsistency from heterogeneous local datasets and varying client computation speeds. Instead of naive weighted aggregation, it applies normalized averaging based on the number of local updates. This drives convergence toward the true stationary point rather than a mismatched objective. By removing solution bias, FedNova delivers more balanced and efficient convergence under heterogeneity.

    \item \textbf{FedPub} [ICML23]. FedPub pioneers a personalized federated learning paradigm for subgraph settings, emphasizing joint improvement of interrelated local GNNs rather than a single global model. To handle subgraph heterogeneity, it uses functional embeddings of local models to compute similarities for weighted server-side aggregation. Each client additionally learns a personalized sparse mask to update only relevant parameter subsets. This prevents collapse of incompatible knowledge and improves personalized graph representations.

    \item \textbf{FedTAD} [IJCAI24]. FedTAD proposes a topology-aware, data-free knowledge distillation strategy to mitigate global GNN degradation under severe subgraph heterogeneity. It decouples node and topology variations, showing how label distribution and structural homophily differences cause class-wise reliability gaps in local knowledge. Based on this, FedTAD dynamically calibrates aggregation to prioritize reliable transfer. This reduces misguided aggregation and improves collaborative training across multi-client subgraphs.

    \item \textbf{FedGTA} [VLDB24]. FedGTA presents a scalable federated graph topology-aware aggregation strategy for collaborative training on large-scale subgraphs. To address slow convergence and heavy optimization overhead, it introduces a personalized mechanism based on topology-aware local smoothing confidence and mixed neighbor features. The method bridges large-scale graph learning and federated optimization without requiring complex local architectures. As a result, FedGTA achieves robust performance and efficient convergence on massive real-world graphs.
    
    \item \textbf{FGGP} [AAAI24]. FGGP introduces a federated graph learning framework with generalizable prototypes to combat performance drops from diverse domain distributions. It decouples the global model into feature extraction and classification levels, connected via semantic prototypes. Using clustered prototypes instead of conventional classifiers, it captures richer domain information and improves class discriminability. It further injects global knowledge through feature-level contrastive learning, enhancing multi-domain prediction and generalization.

    \item \textbf{FGSSL} [IJCAI23]. FGSSL handles severe non-IID federated graph learning by reducing local distortion from node semantics and graph structures. At the node level, it applies contrastive learning to pull local nodes toward same-class global nodes and push away different classes. At the structural level, rather than directly aligning adjacent nodes, it converts adjacency into a similarity distribution and distills relational knowledge from the global model. This preserves structural integrity and discriminability across distributed graphs.

    \item \textbf{DENSE} [NeurIPS22]. DENSE pioneers a practical data-free one-shot federated learning framework that removes dependence on auxiliary public datasets and homogeneous architectures. Within a single communication round, it uses a two-stage pipeline of data generation and model distillation. It transfers only standard model parameters between clients and server while supporting model heterogeneity. This provides an efficient solution for communication-constrained settings with strong collaborative performance.

    \item \textbf{FedCVAE} [ICLR23]. FedCVAE introduces a data-free one-shot federated learning approach using conditional variational autoencoders (CVAEs) to handle extreme statistical heterogeneity. By reframing local learning with a CVAE and shifting its prior distribution, it ensures pipeline security without auxiliary public datasets. Together with FedCVAE-KD, it supports heterogeneous local architectures and efficiently compresses client decoders. It thus retains strong performance in severely non-IID settings with only one communication round.

    \item \textbf{FedSD2C} [NeurIPS24]. FedSD2C is a practical one-shot federated learning framework designed to reduce the severe two-step information loss in traditional data-free knowledge distillation. Rather than aggregating inconsistent local models, it uses a local distiller to synthesize informative distillates from raw client data. By sharing these synthetic distillates with the server, it avoids misleading teacher knowledge under extreme heterogeneity. This direct distillation strategy improves scalability and global model quality in one communication round.

    \item \textbf{FedFisher} [AISTATS24]. FedFisher is a Bayesian one-shot federated learning algorithm that addresses the communication and computational costs of iterative FL. It leverages Fisher information matrices from local models to guide single-round server aggregation. For practical scalability, it uses approximations including diagonal Fisher and K-FAC. This efficient formulation enables an accurate global model while reducing network dependence and privacy-attack susceptibility.

    \item \textbf{GHOST} [ICML25]. GHOST introduces a one-shot federated graph learning framework to overcome catastrophic forgetting and structural bias under restricted communication. Instead of gradual personalized optimization, it builds a proxy model per client to capture localized knowledge. During single-round training, it identifies and consolidates key parameters for critical topological information. This targeted consolidation avoids overfitting to isolated local distributions and improves generalization across diverse graph structures.

    \item \textbf{OASIS} [NeurIPS25]. OASIS pioneers a generative one-shot federated graph learning framework to capture fine-grained structural correlations often missed by image-based generative methods. It introduces a Synergy Graph Synthesizer and Topological Codebook to build a structural latent space and generate informative synthetic graphs. To transfer local structural insight, it uses Wasserstein-driven distillation to capture inter-class semantic affinities and topological relations. This resolves integration challenges and improves robustness in resource-constrained graph learning scenarios.
\end{itemize}

\section{Additional Implementation Details.}\label{implementation}
Experiments are conducted on a Linux platform with NVIDIA GeForce RTX 4090 GPUs and Intel Xeon Gold 6240 CPUs. The implementation uses Python 3.12.7 with PyTorch 2.2.2+cu121 and CUDA 12.1. Beyond the settings reported in the main text, three key hyperparameters are tuned: \(\lambda_{\mathrm{fis}} \in \{10^{-4},10^{-3},10^{-2},10^{-1}\}\), \(\lambda_{\mathrm{sem}} \in \{10^{-4},10^{-3},10^{-2},10^{-1}\}\), and \(\tau \in \{10^{-2},10^{-1},1\}\); the learning rate is selected from \(\{5\times10^{-4},5\times10^{-3}\}\). We use a train/validation/test split ratio of \(0.2/0.4/0.4\). The communication budget is fixed to one round (\(R=1\)), and each selected client performs \(T_L=100\) local update steps per round. On the server side, global proxy-model optimization uses \(T_G=5\) inner steps and \(M=3\) outer iterations during aggregation. For dataset-specific scaling, \(f_{\mathrm{scale}}=10^{-4}\) for Photo and \(f_{\mathrm{scale}}=10^{-7}\) for Ogbn-Arxiv, while \(f_{\mathrm{scale}}=1\) is used for all other datasets.

\section{Ablation Study on Different Numbers of Clients}\label{different number}
To evaluate the scalability and stability of FedHFT, we compare it with 14 baseline methods across four datasets under varying numbers of clients (i.e., 5, 10, and 20 clients). As shown in Table \ref{tab:clients_comparison}, FedHFT consistently achieves superior performance across all settings, demonstrating strong robustness and adaptability against diverse network scales.

\begin{table*}[!h]
\caption{Comparison with state-of-the-art baselines under different client numbers (5, 10, 20) on Cora, CiteSeer, Actor, and Roman-Empire datasets. Green arrows \gup{} indicate improvements over FedAvg, while red arrows \rdown{} denote performance drops relative to FedAvg. The \textbf{best} and \underline{second-best} results in each column are marked in bold and underline, respectively.}
\vspace{1em}
\centering
\renewcommand\arraystretch{1.5}
\resizebox{\textwidth}{!}{
\begin{tabular}{c|l||ccc|ccc} 
\hline\thickhline
\rowcolor{mygray}
\multicolumn{2}{c||}{\textbf{Datasets ($\rightarrow$)}} & \multicolumn{3}{c|}{\textbf{Cora}} & \multicolumn{3}{c}{\textbf{CiteSeer}} \\
\hline
\rowcolor{mygray}
\textbf{Category} & \textbf{Methods ($\downarrow$)} & \textbf{5 Clients} & \textbf{10 Clients} & \textbf{20 Clients} & \textbf{5 Clients} & \textbf{10 Clients} & \textbf{20 Clients} \\
\hline\hline
\multirow{3}{*}{FL} 
& FedAvg [AISTATS17] & 41.27 & 34.09 & 30.46 & 47.21 & 43.00 & 38.13 \\
\cline{2-8}
& FedProx [MLSys20] & 40.16\rdown{1.11} & 33.31\rdown{0.78} & 31.60\gup{1.14} & 44.20\rdown{3.01} & 41.84\rdown{1.16} & 39.17\gup{1.04} \\
& FedNova [NeurIPS20] & 16.25\rdown{25.02} & 15.50\rdown{18.59} & 17.44\rdown{13.02} & 16.48\rdown{30.73} & 16.45\rdown{26.55} & 17.26\rdown{20.87} \\
\hline
\multirow{5}{*}{FGL} 
& FedPub [ICML23] & 36.79\rdown{4.48} & 31.06\rdown{3.03} & 29.88\rdown{0.58} & 35.62\rdown{11.59} & 24.63\rdown{18.37} & 28.45\rdown{9.68} \\
& FedTAD [IJCAI24] & 40.50\rdown{0.77} & 34.62\gup{0.53} & 31.54\gup{1.08} & 44.63\rdown{2.58} & \underline{46.37}\gup{3.37} & 42.21\gup{4.08} \\
& FedGTA [VLDB24] & 13.45\rdown{27.82} & 15.94\rdown{18.15} & 15.93\rdown{14.53} & 16.89\rdown{30.32} & 14.15\rdown{28.85} & 14.97\rdown{23.16} \\
& FGGP [AAAI24] & 40.63\rdown{0.64} & 32.50\rdown{1.59} & 30.74\gup{0.28} & 44.90\rdown{2.31} & 36.85\rdown{6.15} & 36.17\rdown{1.96} \\
& FGSSL [IJCAI23] & 16.34\rdown{24.93} & 13.85\rdown{20.24} & 19.59\rdown{10.87} & 19.01\rdown{28.20} & 19.04\rdown{23.96} & 18.27\rdown{19.86} \\
\hline
\multirow{4}{*}{OFL} 
& DENSE [NeurIPS22] & 34.87\rdown{6.40} & 26.27\rdown{7.82} & 30.15\rdown{0.31} & 30.99\rdown{16.22} & 26.91\rdown{16.09} & 26.84\rdown{11.29} \\
& FedCVAE [ICLR23] & 37.46\rdown{3.81} & 31.91\rdown{2.18} & 30.00\rdown{0.46} & 38.41\rdown{8.80} & 26.91\rdown{16.09} & 24.76\rdown{13.37} \\
& FedSD2C [NeurIPS24] & 44.69\gup{3.42} & 28.84\rdown{5.25} & 34.76\gup{4.30} & 36.18\rdown{11.03} & 30.28\rdown{12.72} & 29.24\rdown{8.89} \\
& FedFisher [AISTATS24] & 40.40\rdown{0.87} & 36.14\gup{2.05} & 30.83\gup{0.37} & 44.14\rdown{3.07} & 34.33\rdown{8.67} & 34.29\rdown{3.84} \\
\hline 
& GHOST [ICML25] & 40.37\rdown{0.90} & 33.55\rdown{0.54} & 26.49\rdown{3.97} & 34.64\rdown{12.57} & 35.45\rdown{7.55} & 28.19\rdown{9.94} \\
& OASIS [NeurIPS25] & \underline{49.52}\gup{8.25} & \underline{40.05}\gup{5.96} & \textbf{45.36}\gup{14.90} & \underline{52.39}\gup{5.18} & 42.82\rdown{0.18} & \textbf{52.88}\gup{14.75} \\
\cline{2-8}
\rowcolor[HTML]{FFF0C1}
\cellcolor{white}\multirow{-3}{*}{OFGL} & \textbf{FedHFT (ours)} & \textbf{62.17}\gup{20.90} & \textbf{52.24}\gup{18.15} & \underline{39.81}\gup{9.35} & \textbf{68.13}\gup{20.92} & \textbf{56.02}\gup{13.02} & \underline{43.42}\gup{5.29} \\

\hline\hline\thickhline
\rowcolor{mygray}
\multicolumn{2}{c||}{\textbf{Datasets ($\rightarrow$)}} & \multicolumn{3}{c|}{\textbf{Actor}} & \multicolumn{3}{c}{\textbf{Roman-Empire}} \\
\hline
\rowcolor{mygray}
\textbf{Category} & \textbf{Methods ($\downarrow$)} & \textbf{5 Clients} & \textbf{10 Clients} & \textbf{20 Clients} & \textbf{5 Clients} & \textbf{10 Clients} & \textbf{20 Clients} \\
\hline\hline
\multirow{3}{*}{FL} 
& FedAvg [AISTATS17] & 23.96 & 24.00 & 25.93 & 21.56 & 21.41 & 18.88 \\
\cline{2-8}
& FedProx [MLSys20] & 23.85\rdown{0.11} & 22.69\rdown{1.31} & 25.03\rdown{0.90} & 21.15\rdown{0.41} & 19.32\rdown{2.09} & 18.45\rdown{0.43} \\
& FedNova [NeurIPS20] & 20.61\rdown{3.35} & 20.01\rdown{3.99} & 20.02\rdown{5.91} & 5.83\rdown{15.73} & 6.64\rdown{14.77} & 6.04\rdown{12.84} \\
\hline
\multirow{5}{*}{FGL} 
& FedPub [ICML23] & 21.51\rdown{2.45} & 24.19\gup{0.19} & 24.06\rdown{1.87} & 19.54\rdown{2.02} & 17.06\rdown{4.35} & 15.36\rdown{3.52} \\
& FedTAD [IJCAI24] & \underline{23.88}\rdown{0.08} & 21.71\rdown{2.29} & \underline{26.54}\gup{0.61} & 21.58\gup{0.02} & 20.34\rdown{1.07} & 19.92\gup{1.04} \\
& FedGTA [VLDB24] & 19.70\rdown{4.26} & 14.20\rdown{9.80} & 21.22\rdown{4.71} & 5.38\rdown{16.18} & 13.29\rdown{8.12} & 5.96\rdown{12.92} \\
& FGGP [AAAI24] & 23.69\rdown{0.27} & 24.37\gup{0.37} & 24.93\rdown{1.00} & 21.83\gup{0.27} & 21.29\rdown{0.12} & 20.61\gup{1.73} \\
& FGSSL [IJCAI23] & 21.64\rdown{2.32} & 19.53\rdown{4.47} & 19.67\rdown{6.26} & 5.97\rdown{15.59} & 6.20\rdown{15.21} & 3.93\rdown{14.95} \\
\hline
\multirow{4}{*}{OFL} 
& DENSE [NeurIPS22] & 23.12\rdown{0.84} & 20.16\rdown{3.84} & 25.60\rdown{0.33} & 14.26\rdown{7.30} & 15.86\rdown{5.55} & 13.89\rdown{4.99} \\
& FedCVAE [ICLR23] & 20.84\rdown{3.12} & 19.92\rdown{4.08} & 24.21\rdown{1.72} & 17.05\rdown{4.51} & 14.93\rdown{6.48} & 12.40\rdown{6.48} \\
& FedSD2C [NeurIPS24] & 23.61\rdown{0.35} & 19.80\rdown{4.20} & 21.49\rdown{4.44} & \underline{24.06}\gup{2.50} & 22.62\gup{1.21} & 22.35\gup{3.47} \\
& FedFisher [AISTATS24] & 19.28\rdown{4.68} & 18.88\rdown{5.12} & 21.22\rdown{4.71} & 17.59\rdown{3.97} & 15.74\rdown{5.67} & 11.63\rdown{7.25} \\
\hline 
& GHOST [ICML25] & 21.86\rdown{2.10} & 23.41\rdown{0.59} & 22.11\rdown{3.82} & 20.35\rdown{1.21} & 19.90\rdown{1.51} & 21.17\gup{2.29} \\
& OASIS [NeurIPS25] & 23.66\rdown{0.30} & \underline{25.51}\gup{1.51} & 23.35\rdown{2.58} & 23.41\gup{1.85} & \underline{23.68}\gup{2.27} & \underline{24.10}\gup{5.22} \\
\cline{2-8}
\rowcolor[HTML]{FFF0C1}
\cellcolor{white}\multirow{-3}{*}{OFGL} & \textbf{FedHFT (ours)} & \textbf{41.54}\gup{17.58} & \textbf{41.51}\gup{17.51} & \textbf{40.87}\gup{14.94} & \textbf{38.35}\gup{16.79} & \textbf{40.82}\gup{19.41} & \textbf{31.95}\gup{13.07}  \\
\hline\hline
\end{tabular}
}
\label{tab:clients_comparison}
\end{table*}

\section{The Performance under various degrees of data heterogeneity}\label{dir}
To evaluate the resilience of FedHFT, we compare it with 14 baseline methods across seven datasets under the non-IID setting with $\alpha=0.1$. As shown in Table~\ref{tab:dir0.1performance}, FedHFT consistently achieves superior performance in all settings, demonstrating strong robustness against data heterogeneity.
\begin{table*}[!h]
\caption{Comparison with state-of-the-art FL, FGL, OFL and OFGL baselines on seven datasets (\(\alpha=0.1\)). We report node classification accuracy (\%) for downstream performance evaluation. Green arrows \gup{} indicate improvements over FedAvg, while red arrows \rdown{} denote performance drops relative to FedAvg. OOM means out-of-memory error. The \textbf{best} and \underline{second-best} results in each column are marked in bold and underline, respectively.}
\vspace{1em}
\centering
\renewcommand\arraystretch{1.5}
\resizebox{\textwidth}{!}{
\begin{tabular}{c|l||ccccccc} 
\hline\thickhline
\rowcolor{mygray}
\textbf{Category} & \textbf{Methods} & \textbf{Cora} & \textbf{CiteSeer} & \textbf{PubMed} & \textbf{Actor} & \textbf{Amz-Photo} & \textbf{Roman-Empire} & \textbf{Ogbn-Arxiv} \\
\hline\hline
\multirow{3}{*}{FL} 
& FedAvg [AISTATS17] & 38.56 & 36.11 & {56.58} & 22.80 & 32.64 & 23.06 & 26.04 \\
\cline{2-9}
& FedProx [MLSys20] & 36.12\rdown{2.44} & \underline{46.59}\gup{10.48} & 40.36\rdown{16.22} & 22.38\rdown{0.42} & 27.88\rdown{4.76} & 22.76\rdown{0.30} & 25.70\rdown{0.34} \\
& FedNova [NeurIPS20] & 15.31\rdown{23.25} & 16.95\rdown{19.16} & 35.69\rdown{20.89} & 19.81\rdown{2.99} & 12.29\rdown{20.35} & 6.54\rdown{16.52} & 3.23\rdown{22.81} \\
\hline
\multirow{5}{*}{FGL} 
& FedPub [ICML23] & 30.77\rdown{7.79} & 28.41\rdown{7.70} & 44.68\rdown{11.90} & 23.14\gup{0.34} & 32.06\rdown{0.58} & 20.82\rdown{2.24} & 20.01\rdown{6.03} \\
& FedTAD [IJCAI24] & 38.26\rdown{0.30} & 40.83\gup{4.72} & 52.27\rdown{4.31} & 21.14\rdown{1.66} & 26.67\rdown{5.97} & 22.31\rdown{0.75} & 15.30\rdown{10.74} \\
& FedGTA [VLDB24] & 12.98\rdown{25.58} & 14.84\rdown{21.27} & 7.41\rdown{49.17} & 19.50\rdown{3.30} & 5.65\rdown{26.99} & 14.23\rdown{8.83} & 9.06\rdown{16.98} \\
& FGGP [AAAI24] & 36.40\rdown{2.16} & 42.74\rdown{6.63} & 47.73\rdown{8.85} & 22.49\rdown{0.31} & 33.62\gup{0.98} & 24.00\gup{0.94} & OOM \\
& FGSSL [IJCAI23] & 13.10\rdown{25.46} & 18.63\rdown{17.48} & 31.86\rdown{24.72} & 19.66\rdown{3.14} & 11.40\rdown{21.24} & 6.65\rdown{16.41} & 2.02\rdown{24.02} \\
\hline
\multirow{4}{*}{OFL} 
& DENSE [NeurIPS22] & 35.93\rdown{2.63} & 23.25\rdown{12.86} & 54.25\rdown{2.33} & 20.68\rdown{2.12} & 19.07\rdown{13.57} & 16.12\rdown{6.94} & 21.78\rdown{4.26} \\
& FedCVAE [ICLR23] & 33.57\rdown{4.99} & 23.30\rdown{12.81} & 46.58\rdown{10.00} & 19.58\rdown{3.22} & 21.61\rdown{11.03} & 13.64\rdown{9.42} & 16.91\rdown{9.13} \\
& FedSD2C [NeurIPS24] & 34.11\rdown{4.45} & 32.74\rdown{3.37} & 47.30\rdown{9.28} & 20.97\rdown{1.83} & 23.96\rdown{8.68} & 26.44\gup{3.38} & \underline{31.22}\gup{5.18} \\
& FedFisher [AISTATS24] & 37.71\rdown{0.85} & 24.42\rdown{11.69} & \underline{57.51}\gup{0.93} & 21.12\rdown{1.68} & {35.15}\gup{2.51} & 22.70\rdown{0.36} & 9.85\rdown{16.19} \\
\hline 
& GHOST [ICML25] & {39.93}\gup{1.37} & 32.55\rdown{3.56} & 53.05\rdown{3.53} & 21.66\rdown{1.14} & 34.56\gup{1.92} & 25.09\gup{2.03} & 20.83\rdown{5.21} \\
& OASIS [NeurIPS25] & \textbf{47.71}\gup{9.15} & 45.00\gup{8.89} & 55.50\rdown{1.08} & \underline{26.08}\gup{3.28} & \underline{51.05}\gup{18.41} & \underline{27.41}\gup{4.35} & 28.36\gup{2.32} \\
\cline{2-9}
\rowcolor[HTML]{FFF0C1}
\cellcolor{white}\multirow{-3}{*}{OFGL} & \textbf{FedHFT (ours)} & \underline{47.60}\gup{9.04} & \textbf{50.92}\gup{14.81} & \textbf{57.58}\gup{1.00} & \textbf{39.32}\gup{16.52} & \textbf{52.09}\gup{19.45} & \textbf{44.33}\gup{21.27} & \textbf{38.52}\gup{12.48}  \\
\hline\hline
\end{tabular}
}
\label{tab:dir0.1performance}
\end{table*}

\section{Privacy Security}\label{privacy}
Privacy security is a critical requirement in federated graph learning \cite{NEURIPS2025_7c818dd4,NEURIPS2025_a924b717,NEURIPS2025_57fbe68c,NEURIPS2025_e6a11b61,NEURIPS2025_adb02256}. To evaluate the privacy robustness of \textit{fedHFT}, we consider a worst-case eavesdropping adversary that can intercept all transmitted client-side updates. Following this threat model, we design two practical reconstruction attempts on \textbf{CiteSeer} and \textbf{Actor} under the setting \texttt{dirichlet\_alpha}=0.05, \texttt{num\_clients}=10.

\textbf{(1) Center-guess attack.}
The attacker does not know the true latent sampling center and therefore tries several guessed centers. Specifically, we evaluate guesses \(\{0,30,60,90\}\), generate attacked pseudo-graphs from intercepted information, and then train attacked models on these pseudo-graphs.

\textbf{(2) Range-overlap attack.}
The attacker further attempts to recover private distributions by enlarging candidate ranges to overlap the hidden latent distribution. We test three ranges: \([{-}100,100]\), \([{-}200,200]\), and \([{-}300,300]\), and repeat the same attacked training/evaluation protocol.

\begin{figure*}[!t] 
    \centering
    
    \begin{minipage}{0.48\textwidth} 
        \centering
        \begin{subfigure}{0.48\linewidth} 
            \centering
            \includegraphics[width=\linewidth]{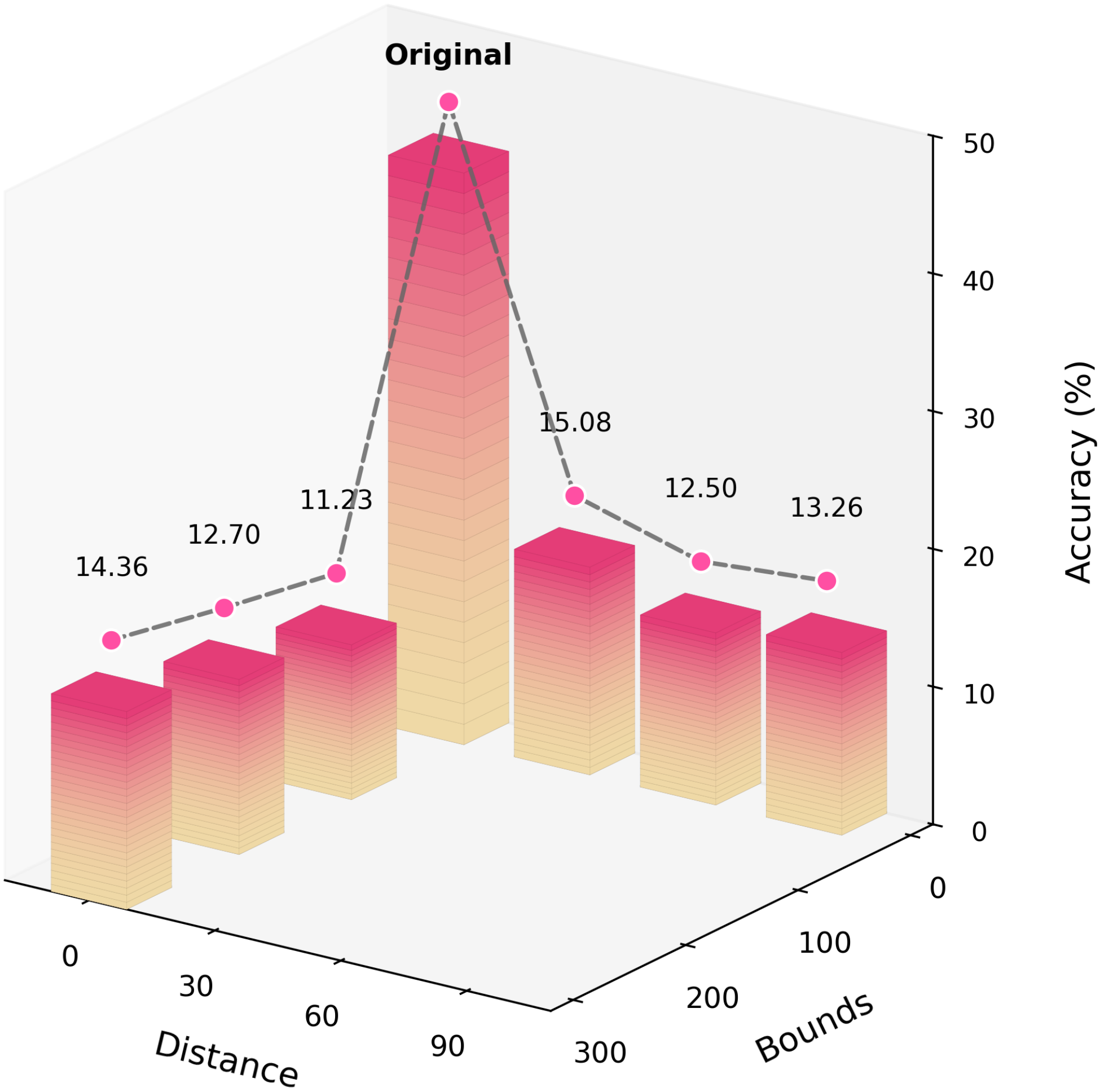}
            \caption{Actor}
            \label{fig:security_center}
        \end{subfigure}
        \hfill
        \begin{subfigure}{0.48\linewidth}
            \centering
            \includegraphics[width=\linewidth]{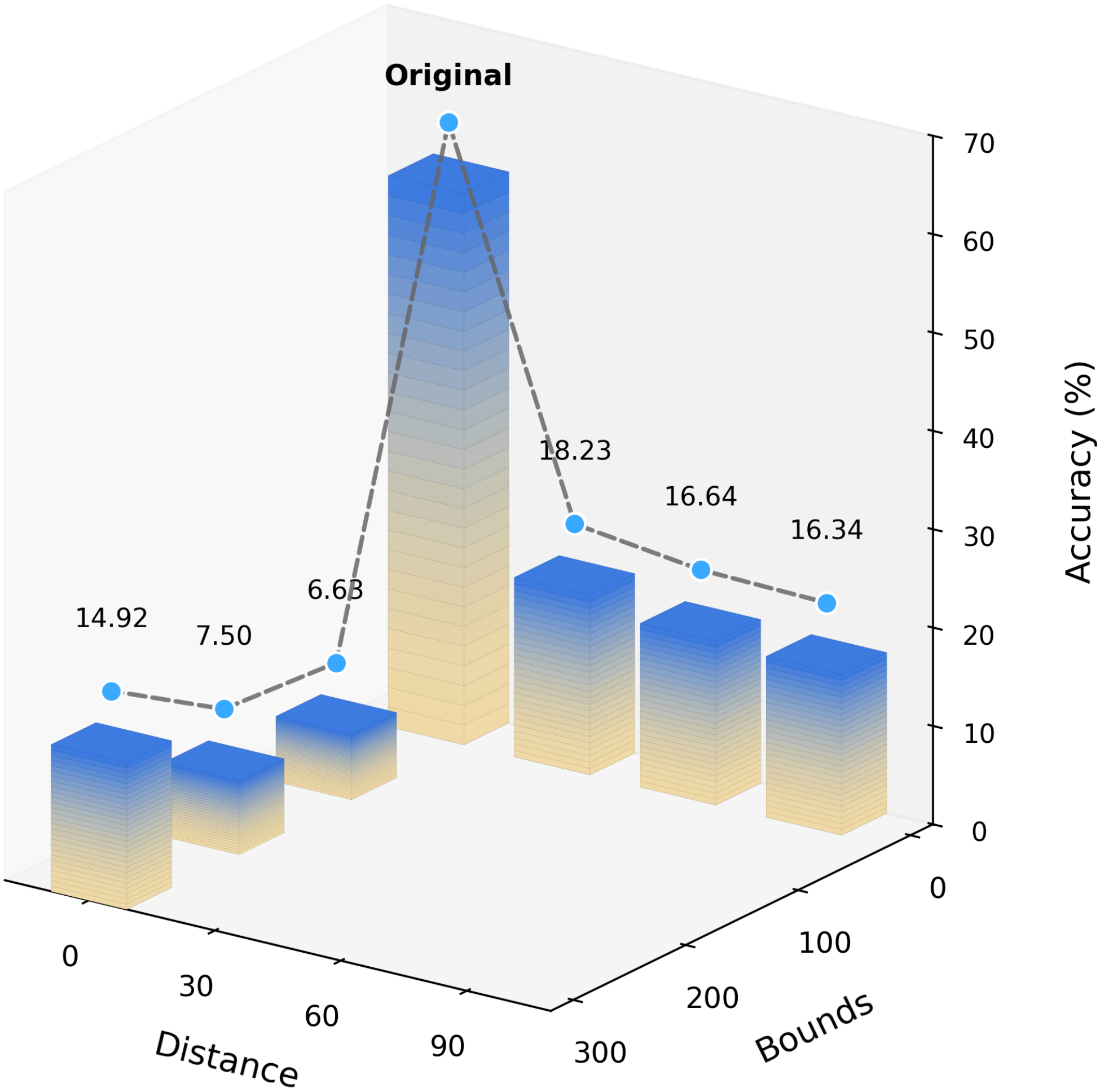}
            \caption{Citeseer}
            \label{fig:security_overlap}
        \end{subfigure}
        \caption{\textbf{Privacy Security Analysis.} Privacy security results on Actor and CiteSeer under center-guess and range-overlap attacks.}
        \label{fig:security}
    \end{minipage}
    \hfill 
    \begin{minipage}{0.48\textwidth} 
        \centering
        \includegraphics[width=\linewidth]{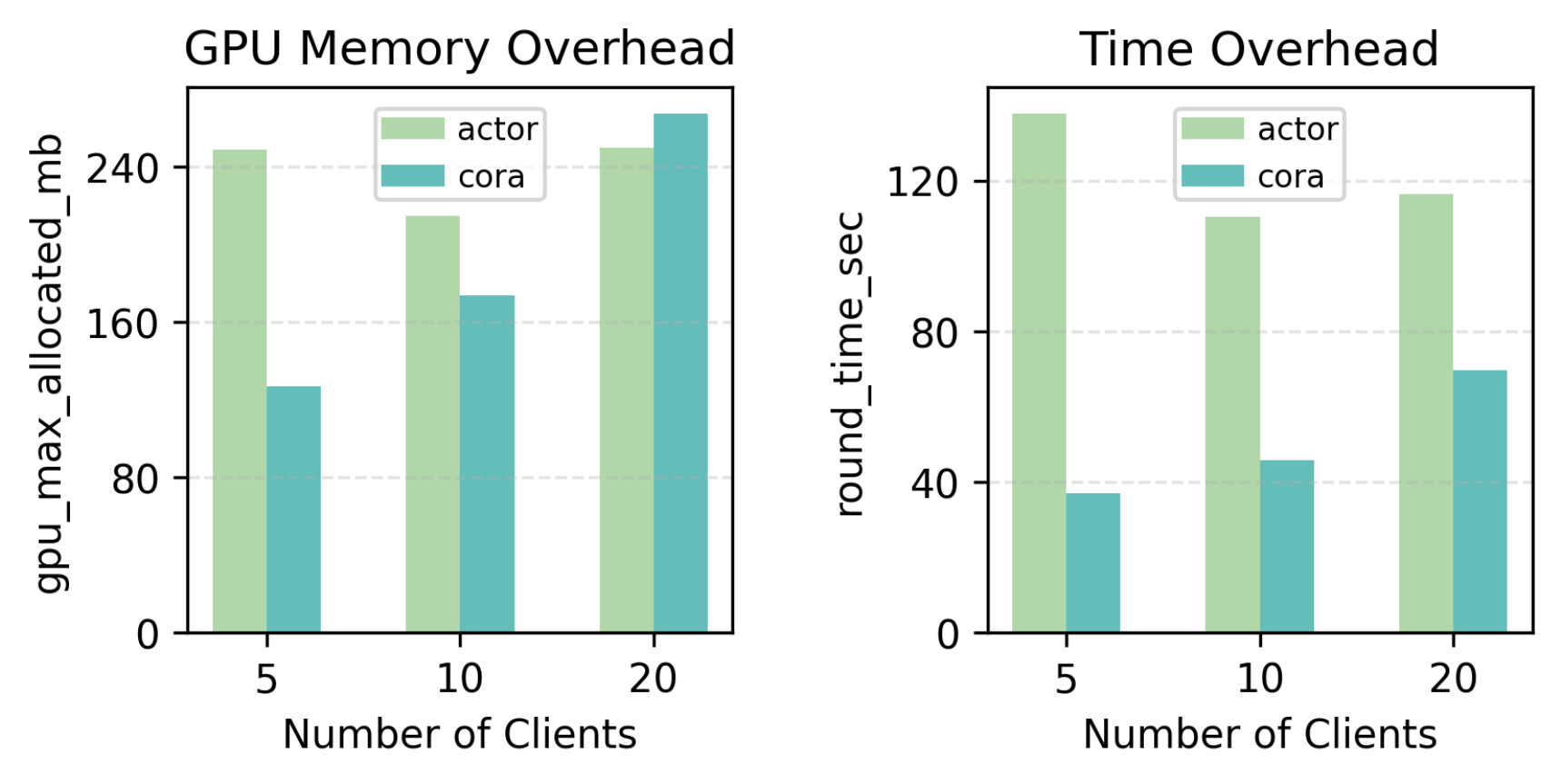} 
        \caption{\textbf{Overhead Analysis.} Comparison of the memory and computation overhead.}
        \label{fig:overhead}
    \end{minipage}

\end{figure*}

Experimental results (Fig.~\ref{fig:security}) show that, in both attack settings, the reconstructed pseudo-graphs consistently yield significantly lower performance than the Original baseline. This gap indicates that intercepted updates are insufficient to recover useful private graph information, demonstrating strong empirical privacy robustness of \textit{fedHFT}.

\section{Equal Communication Budget Analysis}\label{equal_budget}
Our one-round setting follows the standard OFGL evaluation protocol established by GHOST [ICML25] and OASIS [NeurIPS25]. To more directly demonstrate the superiority of FedHFT, we further compare one-round FedHFT with multi-round baselines under matched total communication budgets.

\begin{table*}[!h]
\caption{Comparison under matched total communication budgets: one-round FedHFT versus multi-round baselines on seven datasets. We report node classification accuracy (\%) for downstream performance evaluation. OOM means out-of-memory error. The \textbf{best} and \underline{second-best} results in each column are marked in bold and underline, respectively. On Ogbn-Arxiv, FedAvg, FedProx, and FedTAD are allowed three rounds and FGGP two rounds; all baselines use two rounds on the other datasets; all ten FGGP runs on Ogbn-Arxiv failed with OOM. FedHFT uses one round in all cases.}
\centering
\renewcommand\arraystretch{1.5}
\resizebox{\textwidth}{!}{
\begin{tabular}{l||ccccccc}
\hline\thickhline
\rowcolor{mygray}
\textbf{Methods ($\downarrow$)} & \textbf{Cora} & \textbf{CiteSeer} & \textbf{PubMed} & \textbf{Actor} & \textbf{Amz-Photo} & \textbf{Roman-Empire} & \textbf{Ogbn-Arxiv} \\
\hline\hline
FedAvg [AISTATS17] & 0.3326 & \underline{0.5411} & 0.5366 & \underline{0.2561} & 0.3380 & 0.2190 & \underline{0.2376} \\
FedProx [MLSys20] & 0.3417 & 0.5350 & \underline{0.5397} & 0.2542 & \underline{0.4402} & 0.2183 & 0.2348 \\
FedTAD [IJCAI24] & \underline{0.3468} & 0.5136 & 0.5196 & 0.2534 & 0.3634 & \underline{0.2249} & 0.1011 \\
FGGP [AAAI24] & 0.3318 & 0.4965 & 0.4978 & 0.2489 & 0.3948 & 0.2231 & OOM \\
\rowcolor[HTML]{FFF0C1}
\textbf{FedHFT (ours)} & \textbf{0.5224} & \textbf{0.5602} & \textbf{0.5698} & \textbf{0.4151} & \textbf{0.4931} & \textbf{0.4082} & \textbf{0.4450} \\
\hline\hline
\end{tabular}
}
\label{tab:equal_budget}
\end{table*}

As shown in Table~\ref{tab:equal_budget}, under this stricter matched-budget protocol, one-round FedHFT consistently outperforms every baseline on all seven datasets, further demonstrating its superiority.

\section{overhead}\label{overhead}
\paragraph{Computational Overhead.}
FedHFT is designed for one-shot federated graph learning. Under this constraint, the method intentionally concentrates computation inside the single training cycle to maximize the utility of uploaded knowledge. On the client side, the dominant cost comes from fidelity-driven proxy refinement, including Fisher-guided gradient matching and class-semantic refinement on informative top-$M$ channels, together with topology-aware proxy construction. On the server side, the dominant cost comes from transferability estimation and transferability-steered personalized collaboration. Details are presented in Fig.\ref{fig:overhead}.

\paragraph{Memory Overhead.}
The memory overhead of FedHFT mainly comes from three sources: topology-aware proxy construction and graph-level intermediate tensors, higher-order gradient retention required by Fisher-guided matching, and personalized model states used during server-side transferability-steered aggregation. Relative to lightweight uniform one-shot baselines, memory usage is higher during training. However, this overhead is stage-bounded and scales predictably with client count, feature dimensionality, and local graph size. Since the protocol executes only one communication round, the memory increase is not repeatedly accumulated across rounds. Details are presented in Fig.\ref{fig:overhead}.

\paragraph{Overall Discussion.}
FedHFT addresses the single-transmission bottleneck by coupling fidelity enhancement before upload and transferability-aware collaboration after upload. The framework does not rely on future rounds to repair local distortions or transfer misalignment, and instead enforces knowledge quality and collaboration structure within the only available interaction opportunity. Consequently, FedHFT incurs bounded computational and memory overhead to sharply reduce communication rounds. Importantly, this controlled resource investment is fundamentally justified by the enhanced effectiveness of knowledge transfer. FedHFT effectively translates bounded local computation into highly concentrated knowledge utility, successfully breaking the single-transmission bottleneck as evidenced by an average 15.67\% performance margin over the standard FedAvg paradigm.

\section{Discussion on Limitations} \label{discussion}
While the proposed framework demonstrates strong empirical performance, it exhibits certain limitations. Currently, the experimental validation is primarily constrained to standard benchmark datasets comprising typical homophilic and heterophilic graphs. The generalizability and robustness of the method across more complex data distributions, such as large scale dynamic networks or graphs with severe topological noise, remain to be comprehensively evaluated. Furthermore, the methodology relies on specific hyperparameter configurations. Since the central server lacks direct access to raw local data in a federated environment, optimally tuning these parameters for unseen system deployments presents a practical challenge. 
To address these constraints, future research will explore automated hyperparameter optimization strategies tailored for decentralized environments. Building upon the current high fidelity externalization and transferability steered mechanisms, subsequent work will extend this paradigm to multi round settings and conduct in depth theoretical analyses to facilitate broader real world applications.


\newpage
\section*{NeurIPS Paper Checklist}

\begin{enumerate}

\item {\bf Claims}
    \item[] Question: Do the main claims made in the abstract and introduction accurately reflect the paper's contributions and scope?
    \item[] Answer: \answerYes{} 
    \item[] Justification:The main claims made in the abstract and introduction accurately reflect the contributions and scope of this paper.

\item {\bf Limitations}
    \item[] Question: Does the paper discuss the limitations of the work performed by the authors?
    \item[] Answer: \answerYes{} 
    \item[] Justification: We discuss the limitations in \ref{discussion}

\item {\bf Theory assumptions and proofs}
    \item[] Question: For each theoretical result, does the paper provide the full set of assumptions and a complete (and correct) proof?
    \item[] Answer: \answerNA{}{} 
    \item[] Justification: The paper does not include theoretical results.

    \item {\bf Experimental result reproducibility}
    \item[] Question: Does the paper fully disclose all the information needed to reproduce the main experimental results of the paper to the extent that it affects the main claims and/or conclusions of the paper (regardless of whether the code and data are provided or not)?
    \item[] Answer: \answerYes{}{} 
    \item[] Justification: We fully disclose all the information needed to reproduce the main experimental results in this paper and our code. We are convinced that the obtained results can be reproduced.

\item {\bf Open access to data and code}
    \item[] Question: Does the paper provide open access to the data and code, with sufficient instructions to faithfully reproduce the main experimental results, as described in supplemental material?
    \item[] Answer: \answerYes{}{} 
    \item[] Justification: Code is accessible in this paper.

\item {\bf Experimental setting/details}
    \item[] Question: Does the paper specify all the training and test details (e.g., data splits, hyperparameters, how they were chosen, type of optimizer) necessary to understand the results?
    \item[] Answer: \answerYes{} 
    \item[] Justification:  Details are included in \ref{implementation}

\item {\bf Experiment statistical significance}
    \item[] Question: Does the paper report error bars suitably and correctly defined or other appropriate information about the statistical significance of the experiments?
    \item[] Answer: \answerYes{} 
    \item[] Justification: Statistical significance of the experiments is considered and included in \ref{implementation}

\item {\bf Experiments compute resources}
    \item[] Question: For each experiment, does the paper provide sufficient information on the computer resources (type of compute workers, memory, time of execution) needed to reproduce the experiments?
    \item[] Answer: \answerYes{} 
    \item[] Justification: We provide sufficient information on the computer resources.
    
\item {\bf Code of ethics}
    \item[] Question: Does the research conducted in the paper conform, in every respect, with the NeurIPS Code of Ethics \url{https://neurips.cc/public/EthicsGuidelines}?
    \item[] Answer: \answerYes{} 
    \item[] Justification: The research conducted in the paper strictly adheres to the NeurIPS Code of Ethics, ensuring that all aspects of the work are in compliance with the guidelines

\item {\bf Broader impacts}
    \item[] Question: Does the paper discuss both potential positive societal impacts and negative societal impacts of the work performed?
    \item[] Answer: \answerNA{} 
    \item[] Justification: The research presented in this paper is foundational. It is not directly tied to any specific applications or deployments.
    
\item {\bf Safeguards}
    \item[] Question: Does the paper describe safeguards that have been put in place for responsible release of data or models that have a high risk for misuse (e.g., pre-trained language models, image generators, or scraped datasets)?
    \item[] Answer: \answerNA{} 
    \item[] Justification: This paper poses no such risks

\item {\bf Licenses for existing assets}
    \item[] Question: Are the creators or original owners of assets (e.g., code, data, models), used in the paper, properly credited and are the license and terms of use explicitly mentioned and properly respected?
    \item[] Answer: \answerNA{} 
    \item[] Justification: This paper does not use existing assets.

\item {\bf New assets}
    \item[] Question: Are new assets introduced in the paper well documented and is the documentation provided alongside the assets?
    \item[] Answer: \answerNA{} 
    \item[] Justification: This paper does not release new assets.

\item {\bf Crowdsourcing and research with human subjects}
    \item[] Question: For crowdsourcing experiments and research with human subjects, does the paper include the full text of instructions given to participants and screenshots, if applicable, as well as details about compensation (if any)? 
    \item[] Answer: \answerNA{} 
    \item[] Justification: This paper does not involve crowdsourcing nor research with human subjects.

\item {\bf Institutional review board (IRB) approvals or equivalent for research with human subjects}
    \item[] Question: Does the paper describe potential risks incurred by study participants, whether such risks were disclosed to the subjects, and whether Institutional Review Board (IRB) approvals (or an equivalent approval/review based on the requirements of your country or institution) were obtained?
    \item[] Answer: \answerNA{} 
    \item[] Justification: The paper does not involve crowdsourcing nor research with human subjects.

\item {\bf Declaration of LLM usage}
    \item[] Question: Does the paper describe the usage of LLMs if it is an important, original, or non-standard component of the core methods in this research? Note that if the LLM is used only for writing, editing, or formatting purposes and does \emph{not} impact the core methodology, scientific rigor, or originality of the research, declaration is not required.
    \item[] Answer: \answerNA{} 
    \item[] Justification: The core method development in this research does not involve LLMs as any important, original, or non-standard components.

\end{enumerate}

\end{document}